\documentclass[acmtog,authorversion,screen]{acmart} %

\acmSubmissionID{0194}

\usepackage{subcaption}
\usepackage{array, boldline, rotating}
\usepackage{diagbox}
\usepackage{tikz}
\usepackage{float}
\usepackage{pdfpages}

\usepackage{import}

\newcommand{\red}[1]{{\color{red}{#1}}}
\newcommand{\blue}[1]{{\color{blue}{#1}}}

\newcommand{\NEW}[1]{{\color{black}{#1}}}%
\newcommand{\NEWB}[1]{{\color{black}{#1}}}%

\usepackage{ifpdf}

\usepackage{color}
\definecolor{cyan}{cmyk}{1,0,0,0}
\definecolor{darkgreen}{rgb}{0,0.5,0}
\definecolor{orange}{rgb}{1,0.5,0}
\definecolor{magenta}{cmyk}{0,1,0,0}
\definecolor{darkyellow}{cmyk}{0,0,0.75,0}
\definecolor{gray}{rgb}{0.8,0.8,0.8}

\usepackage{amsmath} %

\usepackage[toc,page]{appendix} %
\usepackage{wrapfig} %
\usepackage{comment}
\usepackage{cuted} %
\usepackage{lipsum} %
\usepackage{mathtools} %
\usepackage{algpseudocode}
\usepackage{gensymb}
\usepackage{float}
\usepackage{soul}
\usepackage{array}
\usepackage{multirow}
\usepackage{hhline}
\usepackage{setspace}
\usepackage{adjustbox}

\algdef{SE}[DOWHILE]{Do}{doWhile}{\algorithmicdo}[1]{\algorithmicwhile\ #1}%

\makeatletter
\renewcommand{\ALG@beginalgorithmic}{\small}
\makeatother

\newcommand{\DELETE}[1]{} %
\newcommand{\IGNORE}[1]{}
\usepackage{datenumber}
\usepackage{calc}
\usepackage[mmddyyyy]{datetime}

\newcounter{datetoday}
\newcounter{diffyears}
\newcounter{diffmonths}
\newcounter{diffdays}

\newcommand{\difftoday}[3]{%
      \setmydatenumber{datetoday}{\the\year}{\the\month}{\the\day}%
      \setmydatenumber{diffdays}{#1}{#2}{#3}%
      \addtocounter{diffdays}{-\thedatetoday}%
      \ifnum\value{diffdays}>0
        \def\diffbefore{}%
        \def\diffafter{left}%
      \else
        \def\diffbefore{}%
        \def\diffafter{ago}%
        \setcounter{diffdays}{-\value{diffdays}}%
      \fi
      \setcounter{diffyears}{\value{diffdays}/365}%
      \setcounter{diffdays}{\value{diffdays}-365*\value{diffyears}}%
      \setcounter{diffmonths}{\value{diffdays}/30}%
      \setcounter{diffdays}{\value{diffdays}-30*\value{diffmonths}}%
      \diffbefore
      \ifnum\value{diffyears}=0
      \else
        \ifnum\value{diffyears}>1
            \thediffyears\space years,
        \else
            \thediffyears\space year,
        \fi
      \fi
      \ifnum\value{diffmonths}=0
      \else
        \ifnum\value{diffmonths}>1
            \thediffmonths\space months
        \else
            \thediffmonths\space month
        \fi
      \fi
      \ifnum\value{diffdays}=0
      \else
        \ifnum\value{diffdays}>1
            \thediffdays\space days
        \else
            \thediffdays\space day
        \fi
      \fi
      \diffafter
}

\def\thickhline{\noalign{\hrule height 1pt}}

\usepackage{booktabs} %

\usepackage[ruled]{algorithm2e} %

\SetAlFnt{\small}
\SetAlCapFnt{\small}
\SetAlCapNameFnt{\small}
\SetAlCapHSkip{0pt}

\acmJournal{TOG}

\setcopyright{rightsretained}
\acmJournal{TOG}
\acmYear{2021}\acmVolume{40}\acmNumber{4}\acmArticle{67}\acmMonth{8} \acmDOI{10.1145/3450626.3459762}

\begin{document}
\title{DeepFormableTag: End-to-end Generation and Recognition of Deformable Fiducial Markers}

\author{Mustafa B. Yaldiz}
\author{Andreas Meuleman}
\author{Hyeonjoong Jang}
\author{Hyunho Ha}
\author{Min H. Kim}
\affiliation{%
  \institution{KAIST}
  \department{School of Computing}
  \city{Daejeon}
  \postcode{34141}
  \country{South Korea}
}
\email{minhkim@kaist.ac.kr}

\renewcommand{\shortauthors}{Yaldiz, Meuleman, Jang, Ha, and Kim}

\begin{abstract}
Fiducial markers have been broadly used to identify objects or embed messages that can be detected by a camera. 
Primarily, existing detection methods assume that markers are printed on ideally planar surfaces. 
The size of a message or identification code is limited by the spatial resolution of binary patterns in a marker. 
Markers often fail to be recognized due to various imaging artifacts of optical/perspective distortion and motion blur. 
To overcome these limitations, we propose \NEW{a novel} deformable fiducial marker system that consists of three main parts:
First, a fiducial marker generator creates a set of free-form color patterns to encode significantly large-scale information in unique visual codes. 
Second, a differentiable image simulator creates a training dataset of photorealistic scene images with the deformed markers, being rendered during optimization in a differentiable manner. 
The rendered images include realistic shading with specular reflection, optical distortion, defocus and motion blur, color alteration, imaging noise, and shape deformation of markers.
Lastly, a trained marker detector seeks the regions of interest and recognizes multiple marker patterns simultaneously via inverse deformation transformation. 
The deformable marker creator and detector networks are jointly optimized via the differentiable photorealistic renderer in an end-to-end manner,
allowing us to robustly recognize a wide range of deformable markers with high accuracy. 
Our deformable marker system is capable of decoding 36-bit messages successfully at $\sim$29 fps with severe shape deformation. 
Results validate that our system significantly outperforms the \NEWB{traditional} 
and data-driven marker methods. 
\NEW{Our learning-based marker system} opens up new interesting applications of fiducial markers, including cost-effective motion capture of the human body, active \NEW{3D} scanning using our fiducial markers' array as structured light patterns, and robust augmented reality rendering of virtual objects on dynamic surfaces.
\end{abstract}




\begin{teaserfigure}
   \hspace{1mm}%
   \includegraphics[width=1\linewidth]{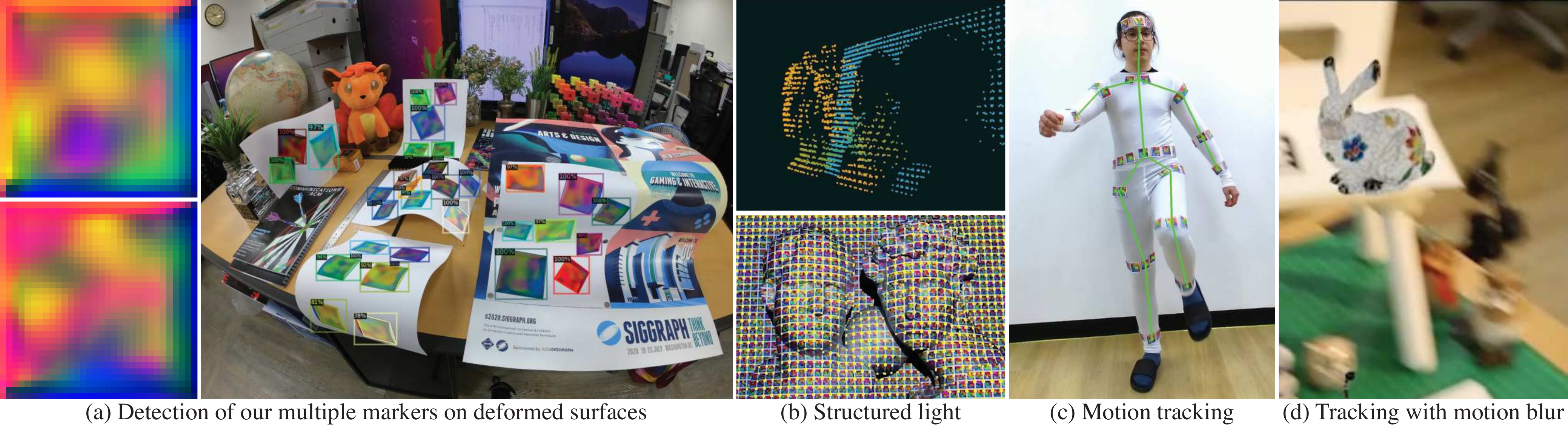}%
   \vspace{-1mm}%
   \caption[]{\label{fig:teaser}%
   (a) shows our learned deformable markers at the leftmost column and presents detection of multiple deformable markers on curved surfaces. All the deformed markers are detected successfully. 
   \NEW{(b) presents the captured results of 3D scanning (top) using a 2D array of multiple fiducial markers as structured light patterns (bottom)}. 
   (c) shows motion tracking of human body motion using our deformable markers.
   (d) demonstrates rendering of a virtual object on the detected markers with strong motion blur.
   	Refer to the supplemental demo video for more results.}%
\end{teaserfigure}

\maketitle

\section{Introduction}
\label{sec:intro}
Fiducial markers with binary patterns, such as QR code~\cite{qrcode:1994:denso}, ARTag~\cite{fiala2005artag}, ARToolKit~\cite{kato1999marker}, AprilTag~\cite{olson2011apriltag}, ArUco~\cite{munoz2012aruco}, and ChArUco~\cite{2020opencv},
have been used broadly to encode identification codes or messages within a square pattern,
enabling object recognition, camera tracking, embedding messages in various computer graphics, vision, and robotics applications.

Despite \NEWB{the} usefulness of these fiducial systems, 
\NEW{they still show several limitations.}
First, all the existing markers assume to be printed on \emph{ideally planar and rigid surfaces}, and thus existing marker systems are incapable of detecting severely deformed marker images.
Second, 
\NEW{marker recognition often fails} 
due to various imaging artifacts of optical/perspective distortion and motion blur when markers are captured in  real-world environments.
Lastly, the size of a message that can be embedded in markers is limited by the spatial resolution of a binary or color pattern.

Even though various hand-crafted visual features have been proposed alternatively in the previous studies~\cite{olson2011apriltag, munoz2012aruco, wang2016apriltag,2020opencv,degol2017chromatag}
to enhance the capability of encoding large-scale information and the performance of detection,
\emph{deformation} of fiducial markers has been rarely discussed due to the hardness of challenges.
Additionally, several assumptions have been made in imaging conditions as well, e.g., there is no optical distortion and motion blur. 
Unfortunately, when capturing \NEW{physically} printed markers with a camera in the real\NEWB{-}world, these assumptions do not always hold, often resulting in unreliable localization, detection, and decoding of marker messages.

Recently, learning-based approaches have been proposed to improve the performance of fiducial marker systems 
\NEWB{through} a set of learnable features~\cite{grinchuk2016learnable,Peace2020E2ETagAE} 
or providing a learning-based enhancement of an existing marker detection method~\cite{hu2019deep}.
However, the former method also train\NEWB{s} the network in an ideal setup where
fiducial markers are located at the center region of training images or perfectly sit on ideal planar surfaces,
missing the important localization capability. 
The latter method detects and refines fiducial corners and the camera pose,
relying on a traditional marker system, \NEWB{ChArUco}~\cite{2020opencv} that assumes ideal imaging scenarios.
To the best of our knowledge, there is no robust and practical solution to tackle the deformed marker detection
problem without sacrificing messaging capability and detection performance.

In this work, we propose 
\NEW{a novel deformable} fiducial marker system specially designed for generating and detecting deformable markers in an end-to-end manner.
Our \NEW{learning-based} method consists of three main parts:
First, a fiducial marker generation network creates learnable fiducial marker patterns to enhance the large-scale message-embedding capability, 
which can be attached \NEWB{to}  real-world free-form surfaces.
Second, a novel differentiable rendering framework creates a realistic training dataset during optimization, where a set of fiducial markers are simulated with surface deformation and realistic imaging conditions, including specular reflection, perspective/lens distortion, defocus and motion blur, color alteration, and imaging noise simulated in a differentiable manner. 
Lastly, a marker detection network consists of a 
localizer network and 
a \NEW{novel} marker decoder network that samples markers' features with respect to deformation. Then, it finally decodes embedded messages within a frame.
Our marker generator and detector are jointly trained by means of the automatically generated \emph{photorealistic} training dataset so that it can detect multiple markers' codes successfully even on severely deformed surfaces under various environments.
Our differentiable rendering framework that  creates immersive markers for real scenes achieves high-quality realism to minimize potential domain gap in the training dataset, outperforming existing learning-based approaches that merely superimpose markers on photographs.

Our end-to-end marker system is capable of creating \NEW{a very large number of messages as unique fiducial markers, for instance, theoretically feasible up to $\sim$68.7 billion (=2$^{36}$) of 36-bit binary messages with a resolution of 32-by-32 pixels}, which can be recognized very robustly and successfully with high bit accuracy at $\sim$29 fps.
Results validate that our fiducial marker system outperforms the traditional and learning-based fiducial marker systems, with robust performance under severely ill-conditioned imaging setup.
Our deformable fiducial marker system opens three interesting applications of fiducial markers.
We demonstrate
(1) cost-effective motion capture of the human body with deformable markers,
(2) single-shot 3D scanning with deformable markers' array used as structured light patterns, and
(3) robust augmented reality rendering of virtual objects on dynamic surfaces.
\NEWB{Figure~\ref{fig:teaser} demonstrates our learnable marker examples and applications using our markers.}
All code, dataset\NEWB{,} and demo are published online to ensure reproducibility (link: \href{https://github.com/KAIST-VCLAB/DeepFormableTag}{https://github.com/KAIST-VCLAB/DeepFormableTag}).

\section{Related Work}
\label{sec:relatedwork}

\begin{figure*}
\centering \footnotesize
    \includegraphics[width=0.98\linewidth]{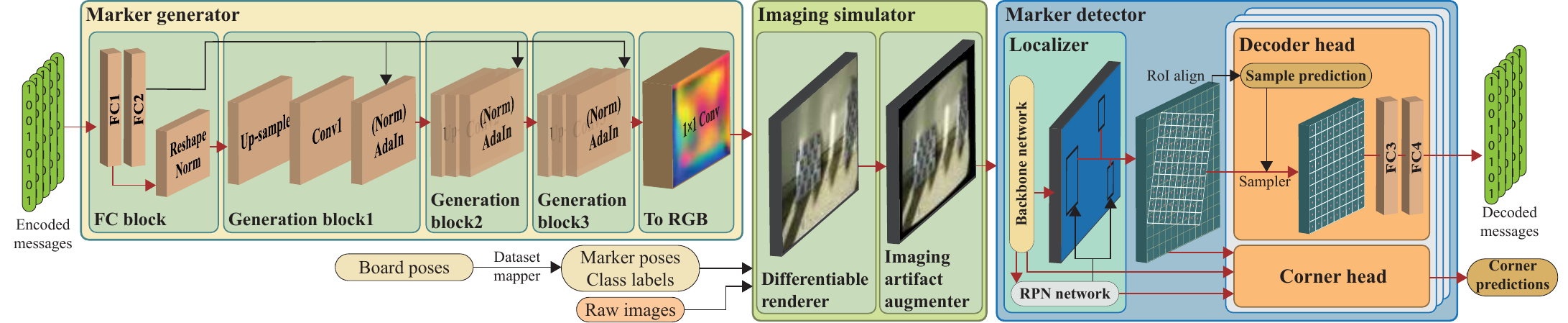}
\caption{\label{fig:model_overview}
Overview. Our end-to-end marker encoding/decoding system consists of three main parts: 
First, the marker generator creates unique fiducial marker patterns from input messages. 
Second, the image simulator renders photorealistic scenes where the markers are simulated with high fidelity through a differentiable rendering framework. The rendered images are also augmented to mimic real-world imaging scenarios. 
Lastly, our marker detector consists of a localizer that proposes regions of interest of each marker, where our decoder network infers a binary message via inverse transformation of deformation.} 
\vspace{-2mm}
\end{figure*}

\paragraph{Traditional markers}
Conventional fiducial marker methods~\cite{fiala2005artag,kato1999marker,olson2011apriltag,munoz2012aruco,2020opencv} consist of two main stages: marker creation and detection.
In the marker generation stage, unique binary identification codes are generated with consideration of detection errors, and maximum inter-marker code distance.
Binary codes in a square shape are used to create payload inside a black border of markers.
The square shape has been preferred to a circular shape \cite{naimark2002circular, bergamasco2011rune, xu2011fourier}
or irregular shape \cite{bencina2005improved}.
\NEWB{This is because four known point correspondences, such as corners, help estimating the marker poses.}
Also, color markers with red and green blocks have been proposed for more efficient detection~\cite{degol2017chromatag}.
In the detection stage, a decoder algorithm detects and recognizes the fiducial markers in an image.
The general approach for the detection algorithms performs quadrilateral detection mainly through an edge detection technique to identify boundaries of the marker.
After the quad-detection process, each marker is warped through an estimated homography, and the payload is decoded with thresholding.

These traditional methods have been developed further with goals to reduce false negatives, to detect smaller and denser tags,
to achieve more robustness against occlusions \cite{wang2016apriltag, krogius2019flexible, garrido2014automatic, romero2018speeded, kallwies2020determining}.
However, several limitations still remain. 
\NEWB{%
First, the amount of encoded information is limited by the spatial resolution of the binary patterns 
\cite{olson2011apriltag, garrido2016generation}, reducing the number of possible markers to several thousands.
Second, the dimensionality of real-world perturbations and actual capturing conditions
are significantly higher than what the current marker methods assume.
For instance, shadow and specular reflection combined with the perspective distortion 
hinder the successful detection of markers.}
\IGNORE{
First, the dimensionality of real-world perturbations is significantly higher than that of the current marker methods.
The amount of encoded information is also limited by the spatial resolution of binary patterns \cite{olson2011apriltag, garrido2016generation}.
Second, in the actual conditions of capturing fiducial markers, markers' patterns are often not recognized successfully.
For instance, perspective distortion occurs depending on the position of the markers in a scene.
It disturbs \NEWB{the} successful detection/recognition of markers.
}
Designed patterns and recognition algorithms often cannot function robustly when imaging conditions are poor.
Recently, \citet{hu2019deep} proposed to train a convolutional neural network, specifically a variant of super-point network~\cite{detone2018superpoint}, to improve the identification capability of the existing ChArUco board marker corners~\cite{2020opencv} and the camera pose in harsh conditions.
They demonstrate the improved detection results, while they only focus on the ChArUco board corner detection as a whole, as opposed to individual ArUco marker detection.
Their corner refinement network requires separate training and needs recollection of a dataset for different marker methods.

\NEW{In addition to square-shaped markers, dot-pattern based markers are developed to detect the deformation of the surfaces 
    \cite{uchiyama2011deformable,narita2016dynamic}.
In order to be detected successfully, each marker needs to occupy a very large portion of object surfaces. 
For \NEWB{these} reasons, detection of \NEWB{a} large number of markers is rather limited, for instance, 
Narita et al.~\shortcite{narita2016dynamic} presented that only two markers at a time are detected as the maximum number.}

\paragraph{Learning-based markers}
\NEW{There have been several efforts to enhance the performance of fiducial marker detection using deep learning.}
An approach of creating learning-based markers was proposed by \citet{grinchuk2016learnable}.
They synthesize markers from binary input codes and render them on random image patches with transformations and distortions. 
They then recognize the decoded message.
This work is a seminal conceptual study that focuses on only creating learnable markers. 
\NEWB{It misses other critical components, for instance, capability of localization, which are required to be used in practice.}
And thus\NEWB{,} these trained markers should be superimposed at the center of images, and this method cannot detect deformed shapes of markers.
There is a concurrent work within this scope. \citet{Peace2020E2ETagAE} attempted to overcome practicability of the existing learnable markers. 
However, the number of their binary codes is limited to a very small number of 30, and the performance is insufficient as 13\,fps for the real-world tracking applications.
As like other existing methods, they assume that fiducial markers are printed on perfectly planar surfaces so that they cannot detect severely deformed fiducial markers.
In contrast, our system is a complete deformable marker method, which is learned with \emph{shape deformation of markers} in an end-to-end manner. 
It can be used in various real-world application scenarios in real time.

\NEW{%
\paragraph{Stegenography}
Steganography and watermarking methods 
have been used to embed hidden messages to an image and to decode them through certain perturbations. 
Recent methods utilize learning-based encoding-decoding of messages with a small perceptual difference in a perturbed image \cite{baluja2017hiding, hayes2017generating, tang2017automatic, wu2018stegnet}. 
HiDDeN~\cite{zhu2018hidden} augments messages by applying color distortions and noise to the encoded image,
assuming perfect alignment during the decoding process without any spatial transformations.
Wengrowski and Dana~\shortcite{wengrowski2019light} capture a dataset to model a camera-display transfer function in order to add distortions to input images. 
However, none of these methods account for the localization problem of encoded messages in input images. 
Recently, Stegastamp~\cite{tancik2020stegastamp} starts to include perspective warp to the perturbation pipeline. 
However, their detection method presents a degraded performance in the real experiments due to the separate architecture of the segmentation network.}

\section{Deformable Fiducial Marker System}
\label{sec:ourmethod}

\subsection{Overview}
\label{sec:overview}
Our deformable fiducial marker system stands on end-to-end training of message encoding and decoding in three folds.
First, our marker generator creates unique marker patterns from input binary messages. 
Second, the created fiducial markers are then supplied to our photorealistic image simulator that consists of a differential renderer and the imaging artifact augmenter. 
The differentiable renderer simulates \NEWB{the} photorealistic appearance of the markers.
The augmenter simulates various imaging artifacts of deformation, perspective/lens distortion, motion blur, 
compression artifacts, and \NEWB{a} variety of illumination conditions.
Lastly, our marker detector is trained with those images. 
It consists of a localizer 
to obtain regions of interest and our marker decoder that detects corners and decodes messages via deformation-aware transformation.
During inference, the identification of markers' binary messages is evaluated with \NEWB{the} dictionary of markers.
Figure~\ref{fig:model_overview} provides an overview.

\subsection{Learnable Marker Generator}
\label{subsec:marker_gen}
\paragraph{Design insight}
In order to create a set of ideal fiducial markers, there are several challenges and trade-offs that need to be resolved. 
Fiducial markers should contain a \emph{rich variety of appearance} to be able to encode and decode \emph{a large number of messages}.
At the same time, a group of fiducial markers should look \emph{similar} among themselves in a way to be robustly detected by a marker detection method. 
On the other hand, from the perspective of the detector, the marker patterns should look \emph{unique} against natural objects' appearance in the real\NEWB{-}world in order to be detected clearly. 
Otherwise, a detector would fail with false\NEWB{-}positive identification.

Traditional fiducial markers consist of black/white 
(or color)
patterns enclosed by a border. 
\NEW{Marker patterns are analytically designed }to achieve the aforementioned goals. 
However, the number of such hand-crafted patterns is very limited to hundreds, and still\NEWB{,} these patterns cannot guarantee robust performances under deformation, distortion, and motion blur.
Recent learning-based approaches are still tested with a limited number of marker patterns, for instance, 30 patterns \cite{Peace2020E2ETagAE}. 

To achieve the capability of embedding a large number of messages, 
we are inspired by steganography, a practice of concealing a message within another message~\cite{johnson1998exploring}.
Our design insight is based on binary encoding and decoding of embedded messages. 
However, different from traditional steganography, our method learns alternative representations of binary codes, which are associated with the latent representations of real-time object detection network jointly trained in the end-to-end manner.
Our approach can embed a very large number of binary messages that grow exponentially by the \NEW{size of the message/number of bits} (two to the power of the number of bits).

\paragraph{Network architecture}
To achieve the strongly distinguishable appearance of a rich variety of marker patterns,
we begin with a baseline generative model, StyleGan \cite{karras2019style}.
The model includes the adaptive instance normalization operator (AdaIn), which learns to align the mean and variance of feature channels.
Our goal is to create visual markers in a certain spatial resolution from binary messages in \NEWB{the} dictionary. 
We found that the AdaIn operator helps  transformation from the binary message domain to the real\NEWB{-}world domain for the markers for better detection and decoding.

We adopt the generation block of StyleGan with the AdaIn operator with a difference. 
The original generative model \NEW{is trained} \NEWB{on} images with progressively higher resolution based on \cite{karras2018progressive}. 
Since we use a relatively small resolution (32$\times$32) compared to natural images, we do not include the progressive resolution approach.

In detail, we apply a fully-connected (FC) linear transformation to the input binary messages. We then normalize these transformed features and apply an activation function. We resize the features from the FC layer into a four-by-four shape. 
As a next, we control the appearance of the markers through convolutional generation blocks. Each generation block applies upsampling at first followed by convolution and normalization operations. 
Style controlling weights of the AdaIn are computed through applying the FC layer transformations  further to the transformed message features. We use the Leaky ReLu activation function after convolutional layers.
The outputs of marker generation networks are finally applied to sigmoid function to normalize the output values within $[0,1]$ range in three color channels for rendering.
Refer to the leftmost yellow block of Figure~\ref{fig:model_overview} for 
\NEWB{an overview of the marker generator network.}

\begin{figure}[pt]
   \centering
   \footnotesize
    \begin{subfigure}{.48\linewidth}
      \frame{\includegraphics[width=\linewidth]{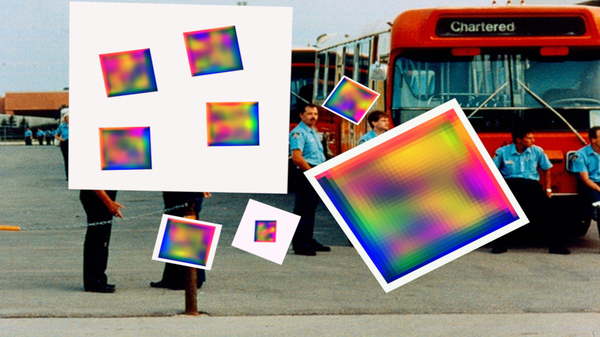}}
      \caption{\label{fig:superimpose_comp} Superimposed rendering}
   \end{subfigure}
   \begin{subfigure}{.48\linewidth}
      \frame{\includegraphics[width=\linewidth]{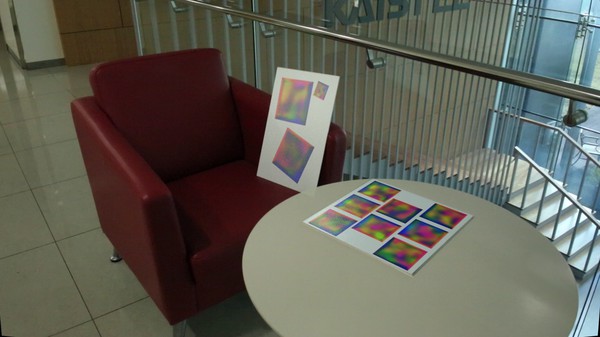}}
      \caption{\label{fig:our_comp_rend} Our photorealistic rendering}
   \end{subfigure}
    \caption[Rendering comparison: superimposition vs our photorealistic rendering]{\label{fig:superimpose-vs-real}%
    (\subref{fig:superimpose_comp})~ An example of randomly superimposed training images with our markers on a COCO dataset, commonly practiced in state-of-the-art methods.
    Original image courtesy of the Mennonite Church USA Archives.
    (\subref{fig:our_comp_rend})~ Our photorealistic rendering of our training dataset.}%
   \vspace{2mm}
  \begin{subfigure}{.32\linewidth}
    \includegraphics[trim={5cm, 4cm, 8.5cm, 2.5cm}, clip, width=.98\linewidth]{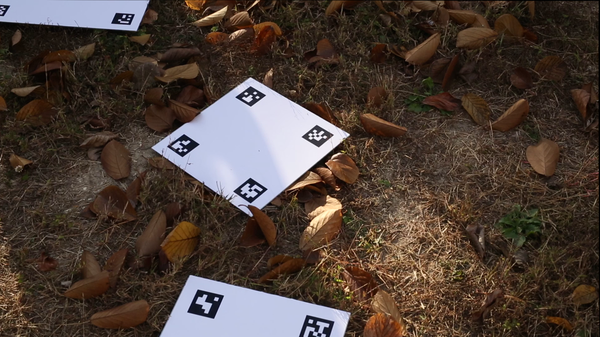}
    \vspace{-1mm}
    \caption{\label{fig:render_raw} Raw}
 \end{subfigure}
 \begin{subfigure}{.32\linewidth}
    \includegraphics[trim={5cm, 4cm, 8.5cm, 2.5cm}, clip, width=.98\linewidth]{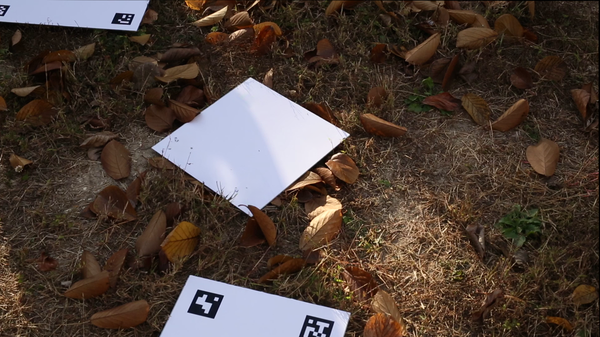}
    \vspace{-1mm}
    \caption{\label{fig:render_preprocessed} Preprocessed}
 \end{subfigure}
  \begin{subfigure}{.255\linewidth}
    \includegraphics[trim={6.6cm, 3.5cm, 8cm, 2.5cm}, clip, width=.98\linewidth]{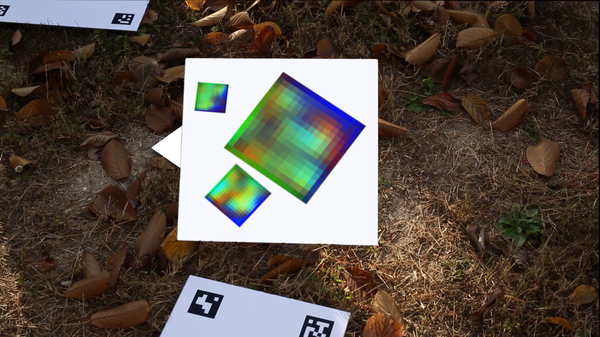}
    \vspace{-1mm}
    \caption{\label{fig:render_superimposed} Markers}
 \end{subfigure}
 \vspace{3mm}
 \begin{subfigure}{.32\linewidth}
    \includegraphics[trim={5cm, 4cm, 8.5cm, 2.5cm}, clip, width=.98\linewidth]{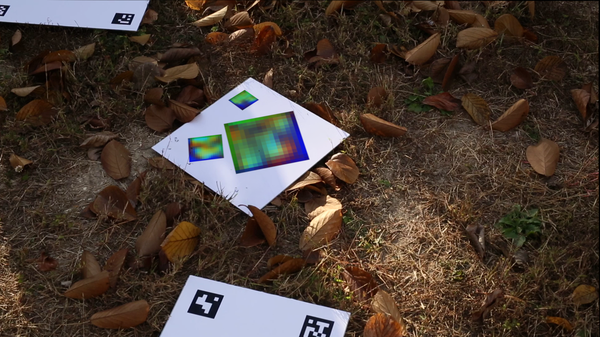}
    \vspace{-1mm}
    \caption{\label{fig:render_diffuse} Diffuse only}
 \end{subfigure}
 \begin{subfigure}{.32\linewidth}
    \includegraphics[trim={5cm, 4cm, 8.5cm, 2.5cm}, clip, width=.98\linewidth]{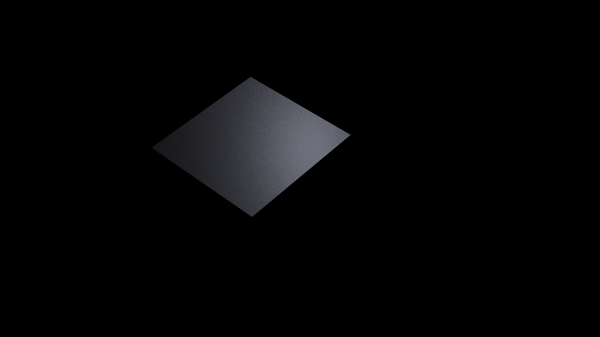}
    \vspace{-1mm}
    \caption{\label{fig:render_specular_only} Specular only}
 \end{subfigure}
  \begin{subfigure}{.32\linewidth}
    \includegraphics[trim={5cm, 4cm, 8.5cm, 2.5cm}, clip, width=.98\linewidth]{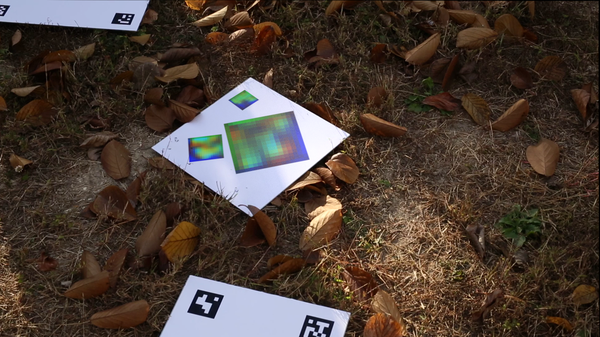}
    \vspace{-1mm}
    \caption{\label{fig:render_ct} Diffuse+Specular}
 \end{subfigure}
 \vspace{-4mm}
 \caption{\label{fig:rendering_comparison} Rendering components. Raw input images~(\subref{fig:render_raw}) are preprocessed by removing initial markers used for the board detection~(\subref{fig:render_preprocessed}). 
(\subref{fig:render_diffuse}) shows diffuse-only rendering of the created marker~(\subref{fig:render_superimposed}).
(\subref{fig:render_specular_only}) presents simulated the specular component of the marker. 
(\subref{fig:render_ct}) shows the final rendering image.}
    \vspace{2mm} 
\centering  \footnotesize
 \begin{subfigure}{.074\textwidth}
    \frame{\includegraphics[width=\linewidth]{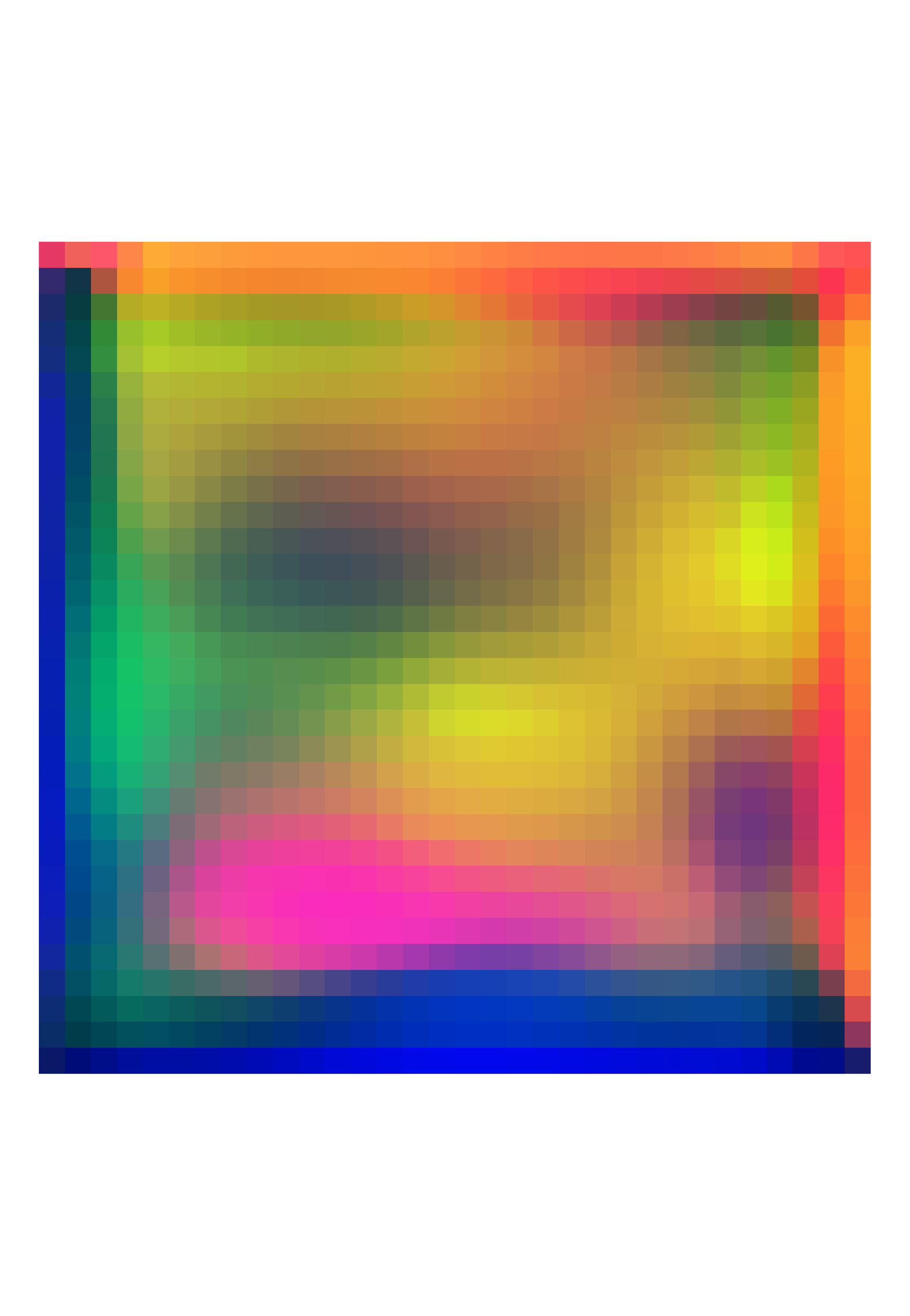}}
    \vspace{-1mm}
 \end{subfigure}
 \begin{subfigure}{.074\textwidth}
    \frame{\includegraphics[width=\linewidth]{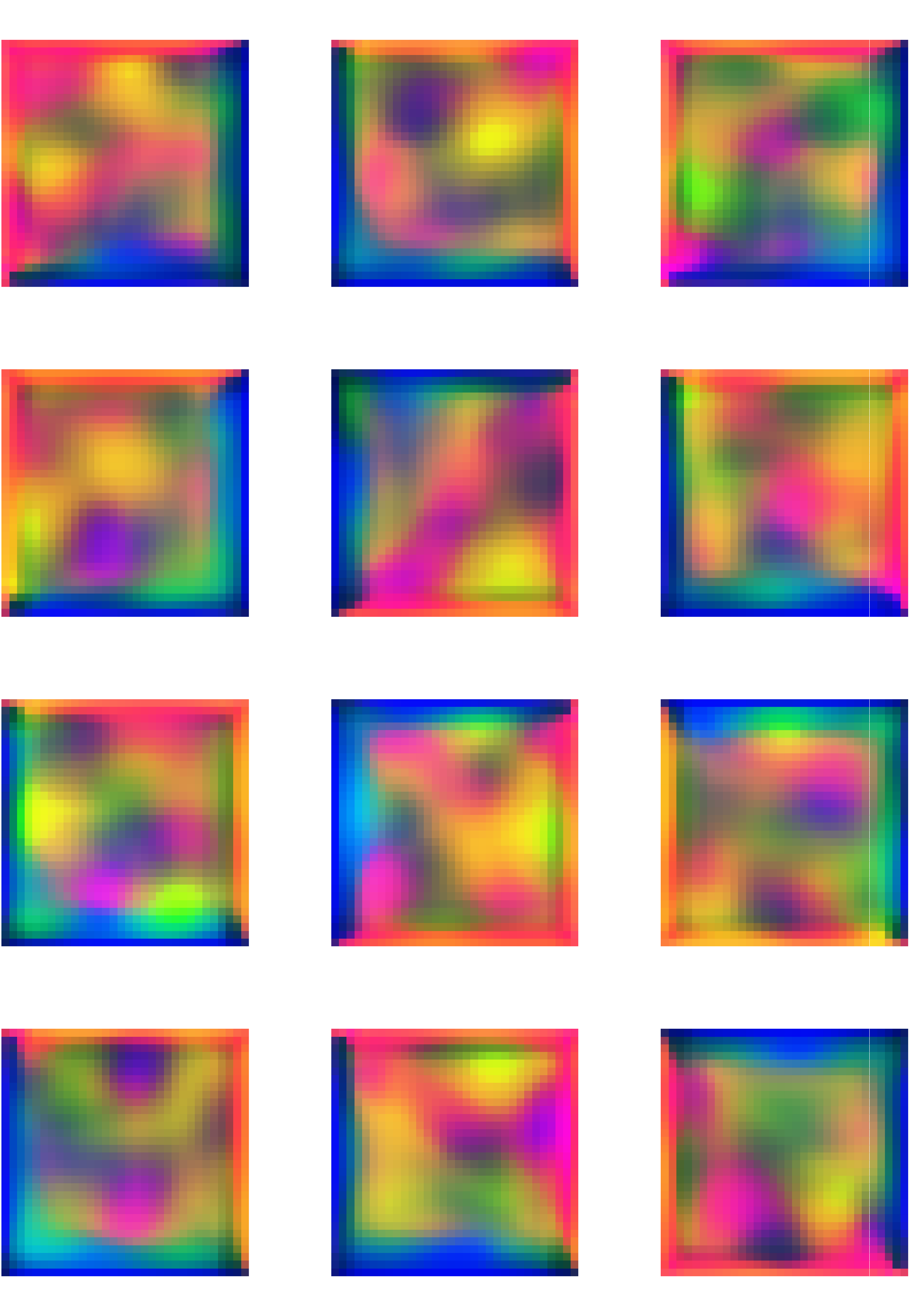}}
    \vspace{-1mm}
 \end{subfigure}
  \begin{subfigure}{.074\textwidth}
    \frame{\includegraphics[width=\linewidth]{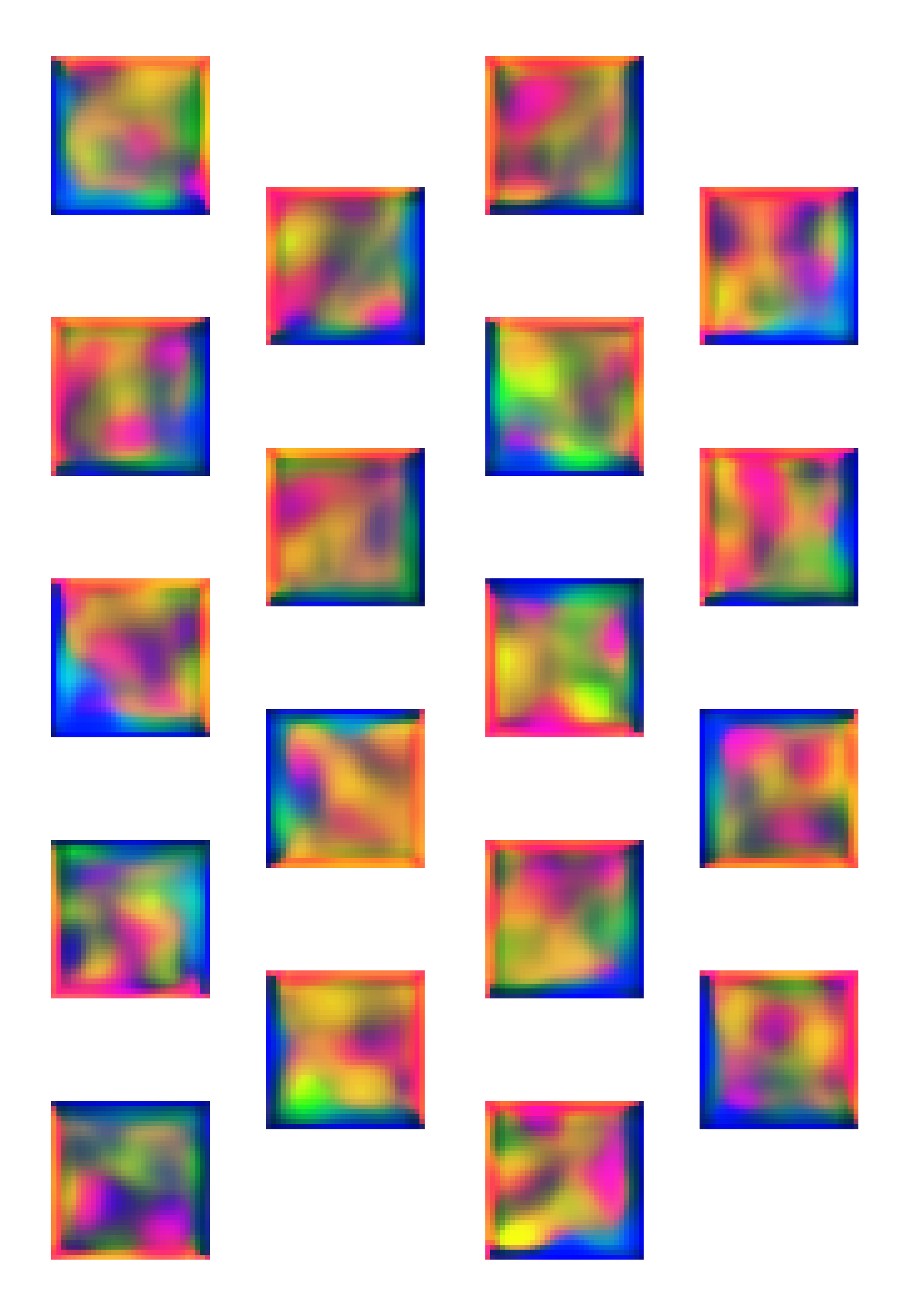}}
    \vspace{-1mm}
 \end{subfigure}
 \vspace{3mm}
 \begin{subfigure}{.074\textwidth}
    \frame{\includegraphics[width=\linewidth]{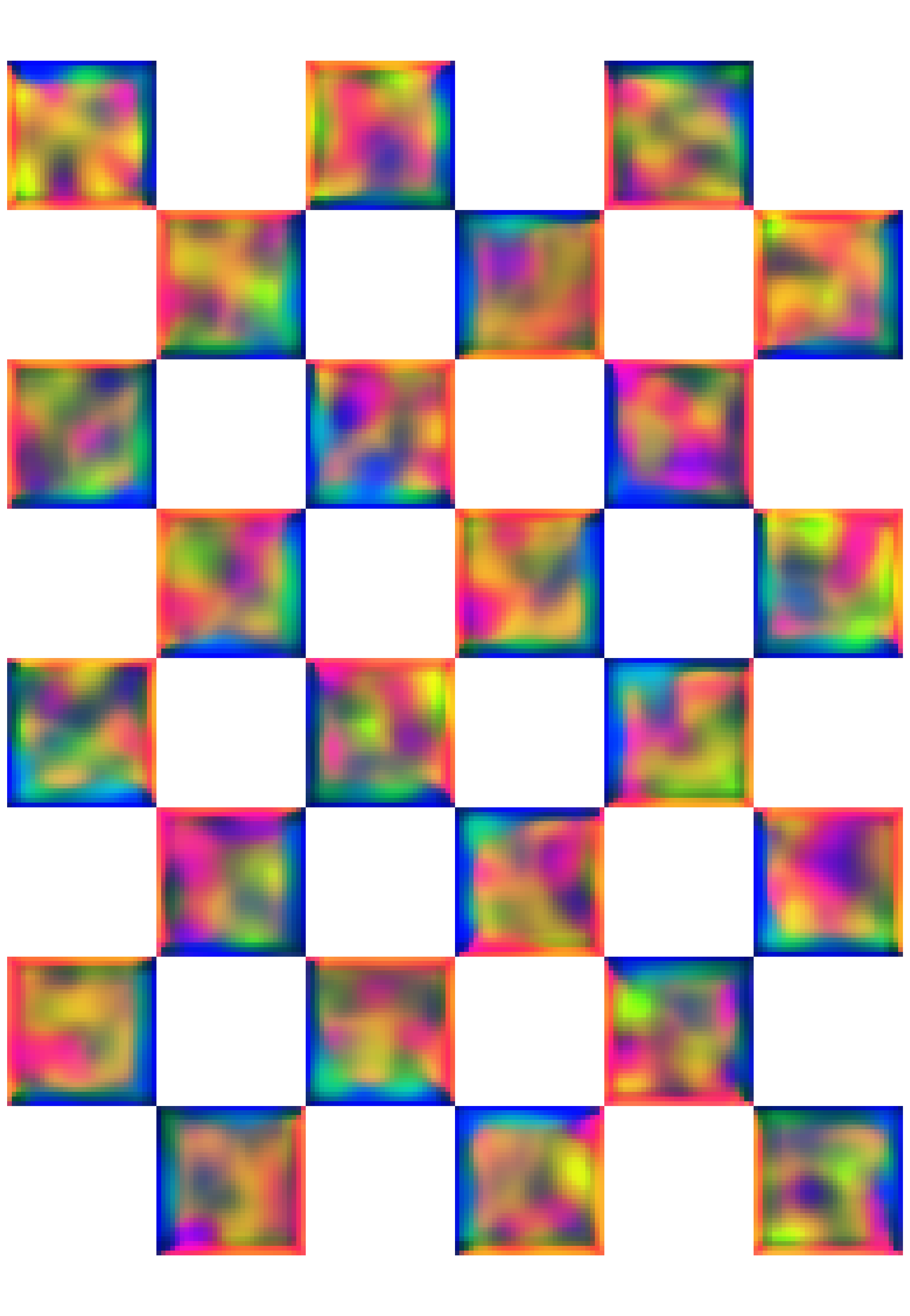}}
    \vspace{-1mm}
 \end{subfigure}
 \begin{subfigure}{.074\textwidth}
    \frame{\includegraphics[width=\linewidth]{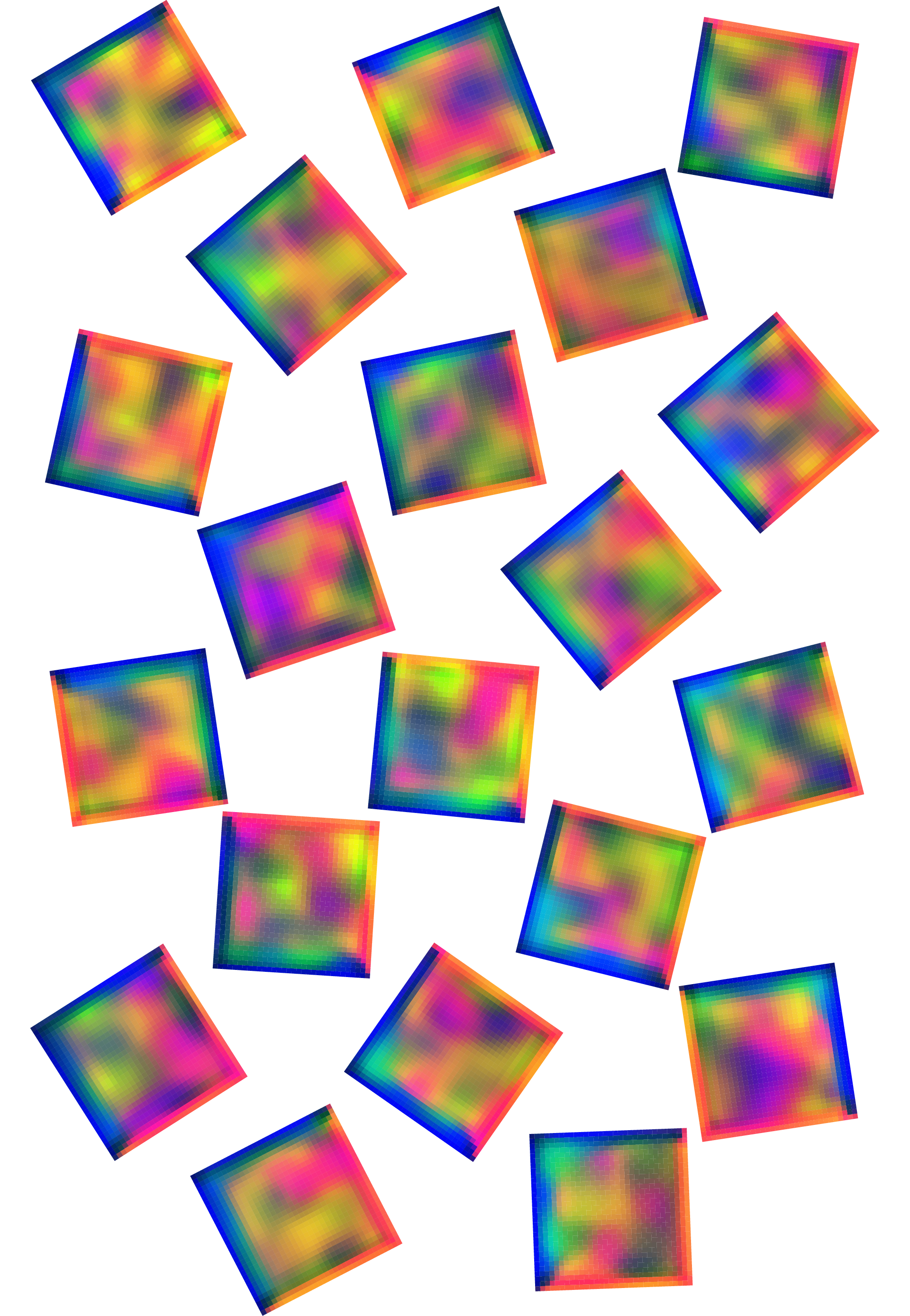}}
    \vspace{-1mm}
 \end{subfigure}
  \begin{subfigure}{.074\textwidth}
    \frame{\includegraphics[width=\linewidth]{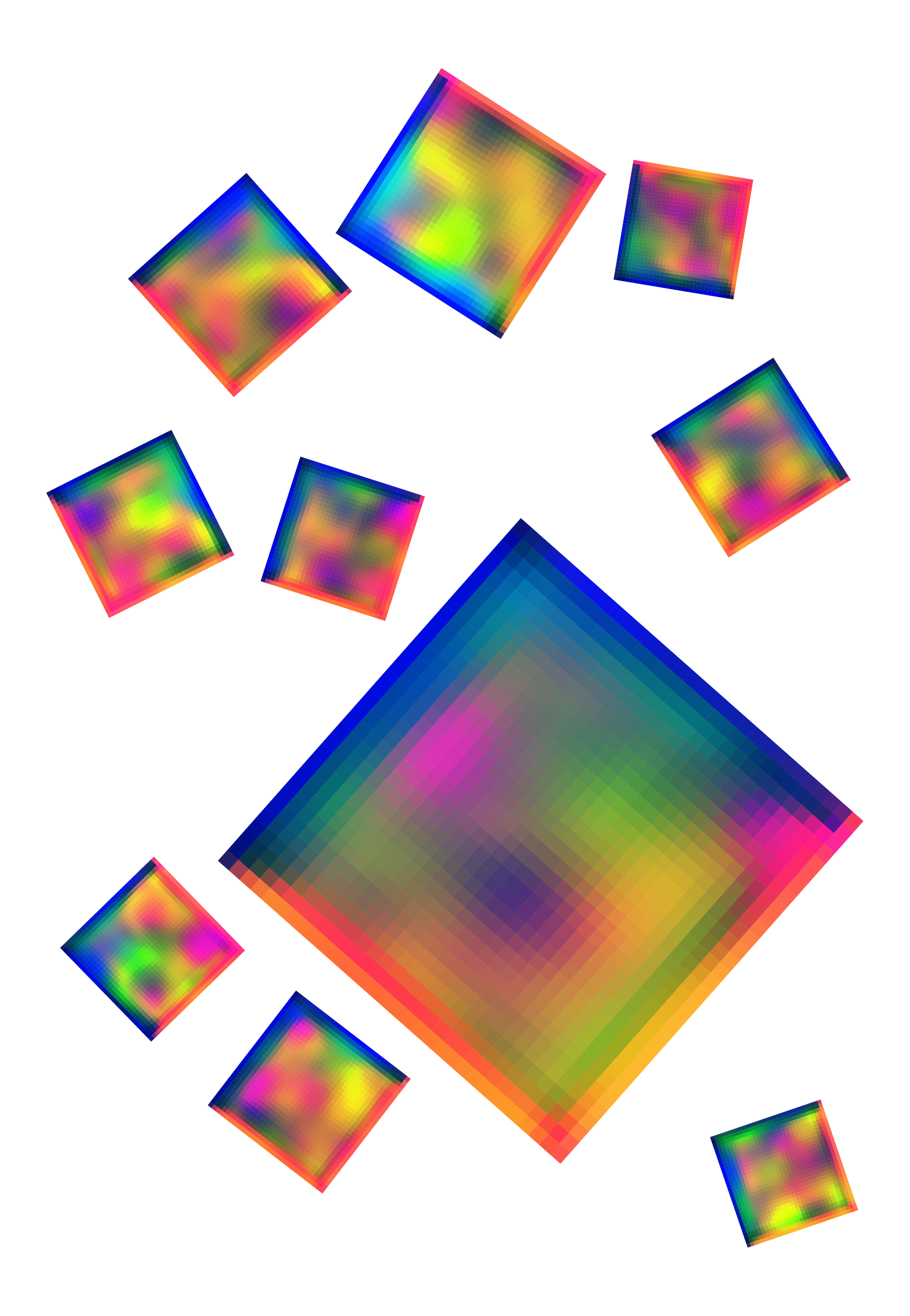}}
    \vspace{-1mm}
 \end{subfigure}
 \vspace{-5mm}
 \caption{\label{fig:marker_configurations} Marker configuration examples. We place markers on the boards with various alignments during training.}
\end{figure}

\subsection{Differentiable Image Simulator}
\label{sec:our_method}

\paragraph{Design insight}
Our differentiable image simulator is designed with two insights.
First, the existing learning-based approaches \cite{hu2019deep,Peace2020E2ETagAE} have trained the marker generator or detector by superimposing markers on photographs from the MS COCO  \cite{lin2014microsoft} or ImageNet \cite{deng2009imagenet} datasets. 
Markers are placed at random positions, ignoring scene geometry and illumination. 
See Figure~\ref{fig:superimpose-vs-real} for an example.
The realism of the superimposed training dataset is far lower than actual images that include fiducial markers in the real-world scenes. 
The domain gap between the randomly superimposed images and natural images of fiducial markers could degrade training performance severely. It requires further efforts of domain adaptation. It is verified later through an ablation study in Table~\ref{tab:rendered_main}. 
To mitigate the domain gap in training, our approach is to create  \emph{photorealistic rendering images} for training, where the generated markers are seamlessly rendered by physically-based rendering composed in real photographs.
Second, our method has increased the messaging capability in an exponential scale by means of color patterns. 
However, it becomes more critical to learn the appearance variation of color patterns in the real-world environment
because a color code might be \NEW{misinterpreted as} 
a different code depending
on environments. 
To mitigate these two challenges, we devise a differentiable image simulator for \emph{photorealistic} rendering of the generated fiducial markers.

\subsubsection{Differentiable Rendering}
\label{sec:rendering}

In our end-to-end optimization framework, we need a large number of photorealistic training images,
where the generated markers are placed in real-world scenes at runtime.
If we render the generated fiducial markers through a physically-based rendering framework, the rendering process may take a several hours to obtain a single high-resolution image. 
Instead, we introduce a practical solution to create realistic marker images efficiently.
In brief, we first prepare thousands of \NEWB{real-world video frames} 
under different illumination, 
where multiple white boards as space holder are placed. Then, we render the generated markers on the board with diffuse shading of scene illumination and synthetic specular reflection. 
Lastly, our augmentation operators simulate various imaging artifacts of deformation, projection, noise, illumination, etc.

\paragraph{Preprocessing and placement}
\label{sec:rendering-mapping}
For our main training pipeline, we collect a real-image dataset, so-called \emph{placement dataset}. 
It consists videos of planar white boards in real scenes from simple to complex scenes with many objects (see Figure \ref{fig:render_raw}).  
We capture video frames of 140 real-world scenes under various illumination conditions indoor and outdoor with two DSLR cameras (Canon EOS 5Ds equipped with 22mm and 50mm lenses, respectively).
To capture high-quality photographs without imaging artifacts, such as motion blur, we capture videos at a short shutter rate using two 3-axis gimbals.
Note that the white boards include the initial Aruco markers \cite{munoz2012aruco} on the corners, 
which are used for automatic localization of the white board of space holder. 
We detect the white board's location and orientation from the initial markers,
\NEW{and estimate its homography for the marker placement later on.}
\NEW{To provide a spotless surface for rendering, we remove the initial Aruco markers 
using two inpainting algorithms \cite{liu2018image, yu2018generative} (see Figure \ref{fig:render_preprocessed}). 

During training, we create 96 unique messages at each iteration, and feed them to the generator. 
For each board, we select one of the preset layout configurations (as shown in Figure~\ref{fig:marker_configurations}). 
Next, we place markers by randomly selecting from 96 markers according to \NEWB{the} boards homography and layout. 
Using 96 unique markers per iteration provided better convergence, as opposed to using unique markers for each location.
In addition, identification indices of the markers later provide us ground-truth labels to calculate the average precision score.}

\paragraph{Rendering of markers}
For \NEWB{the} planar board, we use diffuse white paper evenly attached on an aluminum plate
so that surface exhibits mostly Lambertian reflection. 
Assuming paper surface $m_p$ and marker's surface $m_t$ have constant BRDFs: $f_{\mathbf{x},\vec{n},m_p}(\vec{w},\vec{v})=\frac{\rho_p}{\pi}$, $f_{\mathbf{x},\vec{n}, m_t}(\vec{w},\vec{v})=\frac{\rho_t}{\pi}$, for the camera direction $\vec{v}$, radiance from surface point $\mathbf{x}$ in 3D is: $L(\mathbf{x}, \vec{v}) = \frac{\rho}{\pi}\int_{\Omega}{L(\vec{w},\mathbf{x})cos(\theta)d\omega}$. 
Therefore, the reflected radiance of the surface is proportional to the diffuse reflectance ratio between the target rendering surface and the paper surface. 
The reflected radiance can be computed as:
$L_{m_t}(\mathbf{x}, \vec{v}) = L_{m_p}(\mathbf{x}, \vec{v}) \frac{\rho_t}{\rho_p}$.
The marker colors are then multiplied by corresponding surface pixels and divided by the surface's proportional diffuse reflectance
(Figure~\ref{fig:render_diffuse}).
We calculate color in linear color space, then gamma correct it before providing it to the detection network.

Real-world paper and printer ink often present specular reflection\NEWB{s}. 
To simulate shininess of real-world paper, 
we introduce specular rendering (Figure~\ref{fig:render_specular_only}). We use \NEW{the} surface normals, viewing directions, and surface pixel values to create specular effects. 
We select \NEWB{the} light direction from perfect reflection direction of the brightest point within 
\NEW{the board}. 
And we define the light's color as the normalized color of the brightest point, with a randomized intensity power value, 
scaled by \NEWB{the} boards overall brightness.  
We use the Cook-Torrance microfacet BRDF model with the GGX microfacet distribution function  \cite{walter2007microfacet} to render specular highlight on surfaces.
For the specular component, we fix roughness and specular albedo across the surface with small normal perturbations. 
This approach shows viable renderings of specular effects \NEWB{that} occur in the wild (see Figures~\ref{fig:superimpose-vs-real}b and~\ref{fig:render_ct} for rendering results used for training).

\begin{figure}[t]
 \centering  \footnotesize
 \begin{subfigure}{.125\textwidth}
    \includegraphics[trim={5cm, 4cm, 7.5cm, 2cm}, clip, width=.97\linewidth]{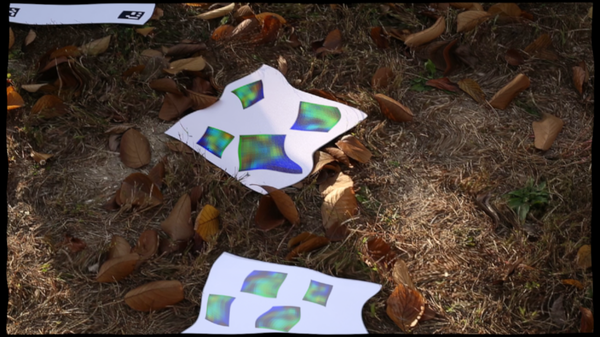}
    \vspace{-2mm}
    \caption{\label{fig:aug1} Deformation}
 \end{subfigure}
  \begin{subfigure}{.105\textwidth}
    \includegraphics[trim={8cm, 5cm, 9cm, 3.5cm}, clip, width=.97\linewidth]{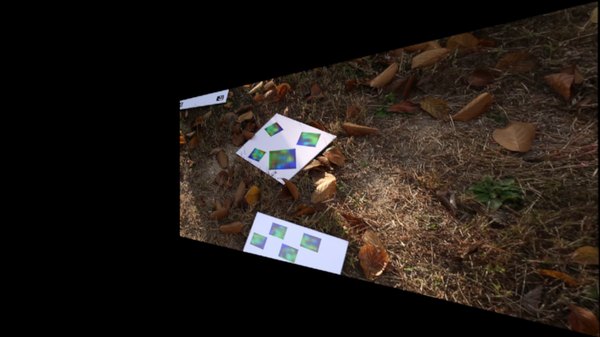}
    \vspace{-2mm}
    \caption{\label{fig:aug2} Perspective }
 \end{subfigure}
 \begin{subfigure}{.103\textwidth}
    \includegraphics[trim={6cm, 4cm, 8cm, 2cm}, clip, width=.97\linewidth]{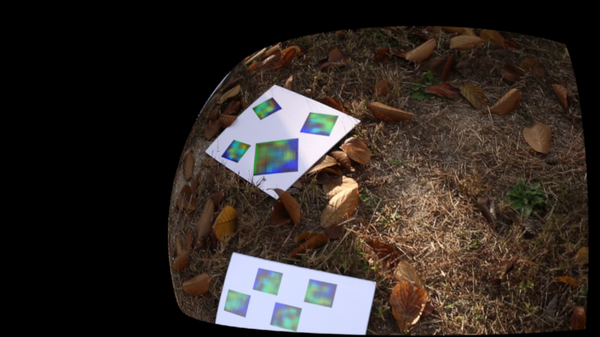}
    \vspace{-2mm}
    \caption{\label{fig:aug3} Radial dist.}
 \end{subfigure}
 \begin{subfigure}{.125\textwidth}
    \includegraphics[trim={5cm, 4cm, 7.5cm, 2cm}, clip, width=.97\linewidth]{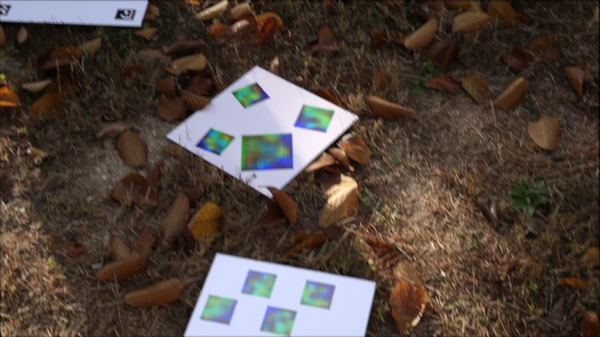}
    \vspace{-2mm}
    \caption{\label{fig:aug4} Motion blur}
 \end{subfigure}
  \begin{subfigure}{.115\textwidth}
    \includegraphics[trim={5cm, 4cm, 7.5cm, 2cm}, clip, width=.97\linewidth]{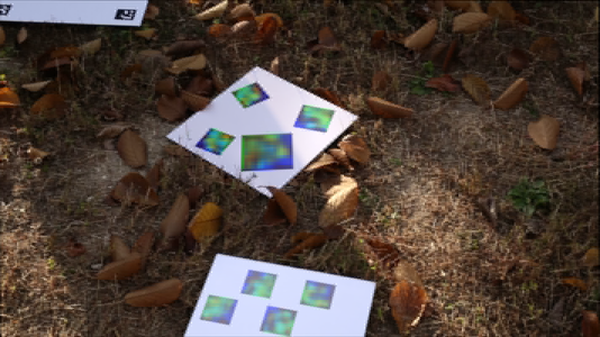}
    \vspace{-2mm}
    \caption{\label{fig:aug5} Jpeg}
 \end{subfigure}
 \vspace{-5mm}%
 \begin{subfigure}{.115\textwidth}
    \includegraphics[trim={5cm, 4cm, 7.5cm, 2cm}, clip, width=.97\linewidth]{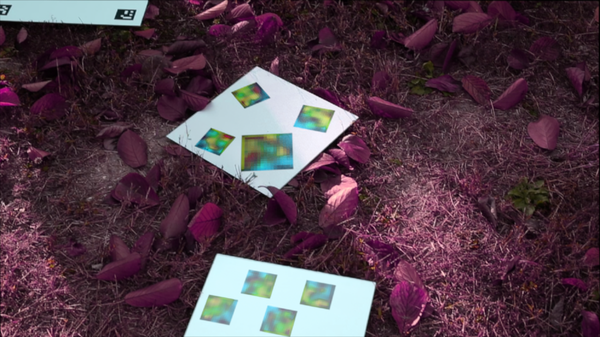}
    \vspace{-2mm}
    \caption{\label{fig:aug6} Color}
 \end{subfigure}
 \begin{subfigure}{.115\textwidth}
    \includegraphics[trim={5cm, 4cm, 7.5cm, 2cm}, clip, width=.97\linewidth]{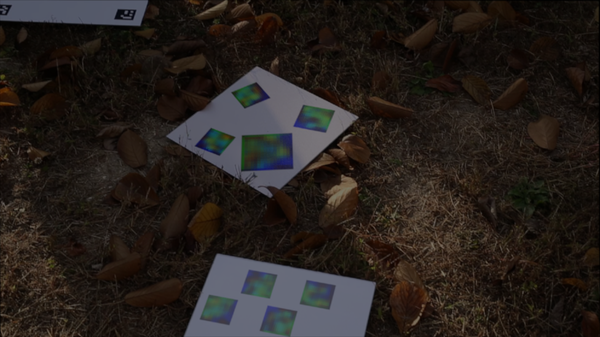}
   \vspace{-2mm}
    \caption{\label{fig:aug7} Brightness}
 \end{subfigure}
  \begin{subfigure}{.115\textwidth}
    \includegraphics[trim={5cm, 4cm, 7.5cm, 2cm}, clip, width=.97\linewidth]{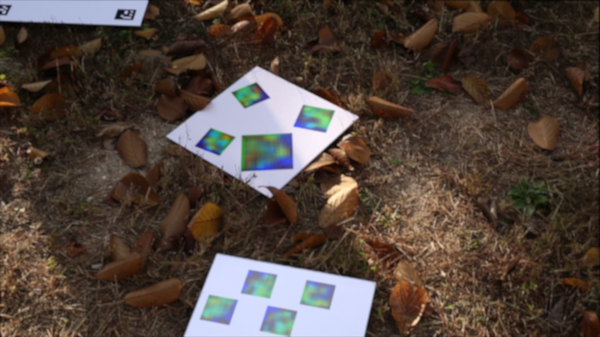}
   \vspace{-2mm}
    \caption{\label{fig:aug8} Defocus blur}
 \end{subfigure}
 \vspace{5mm}
 \caption{\label{fig:augmentations}
     Examples of imaging artifacts. Deformation, perspective and radial distortions provide geometric augmentations, while the rest provides color-based augmentations of motion blur, JPEG compression artifacts, illumination color change, scene brightness change, defocus artifacts, etc.
 }
 \vspace{-2mm}
\end{figure}

\subsubsection{Simulation of imaging artifacts}
\label{sec:augmentation}
As discussed earlier, many edge cases exist for recognition. Hence, it is not enough to simply feed rendered images with labels to the detection pipelines. 
Therefore, we devise an imaging artifact simulator for differentiable augmentations of rendered images to generalize for the difficult conditions.
	\NEWB{Figure~\ref{fig:augmentations} presents examples of artifacts.}

\paragraph{Deformation}
To simulate scenarios of deformed and non-flat placement of markers, we warp the rendered image using \emph{thin-plate splines (TPS)}~\cite{duchon1977tps} at training time. 
\NEW{We intentionally deform the surfaces in an algorithmic manner, rather than capturing predefined non-planar surfaces. 
This decision 
\NEW{is taken to avoid any}
potential risk of the network's overfitting to a limited number of predefined shapes.
Also, preparing real objects of various shapes is cumbersome.}

In detail, we first define control points in the target image as a uniform grid. 
We then shift them, each in a random direction and amplitude, to obtain the control points in the source image. 
\NEW{Finally, we used TPS} to create a dense sampling grid following the control points.
This process allows for varied deformation and distortion while the properties of TPS ensure smoothness. 
See Figure~\ref{fig:aug1} for an example.

It is worth noting that while we apply these non-linear geometric distortions, we had to recalculate ground-truth labels such as marker corners. 
We also store and recalculate uniform marker sampling locations, 
which \NEW{are later} used to teach \NEWB{the} network to explicitly invert all the nonlinear transformations applied to the marker.

\begin{figure}[pt]
  \centering
   \includegraphics[width=1.0\linewidth]{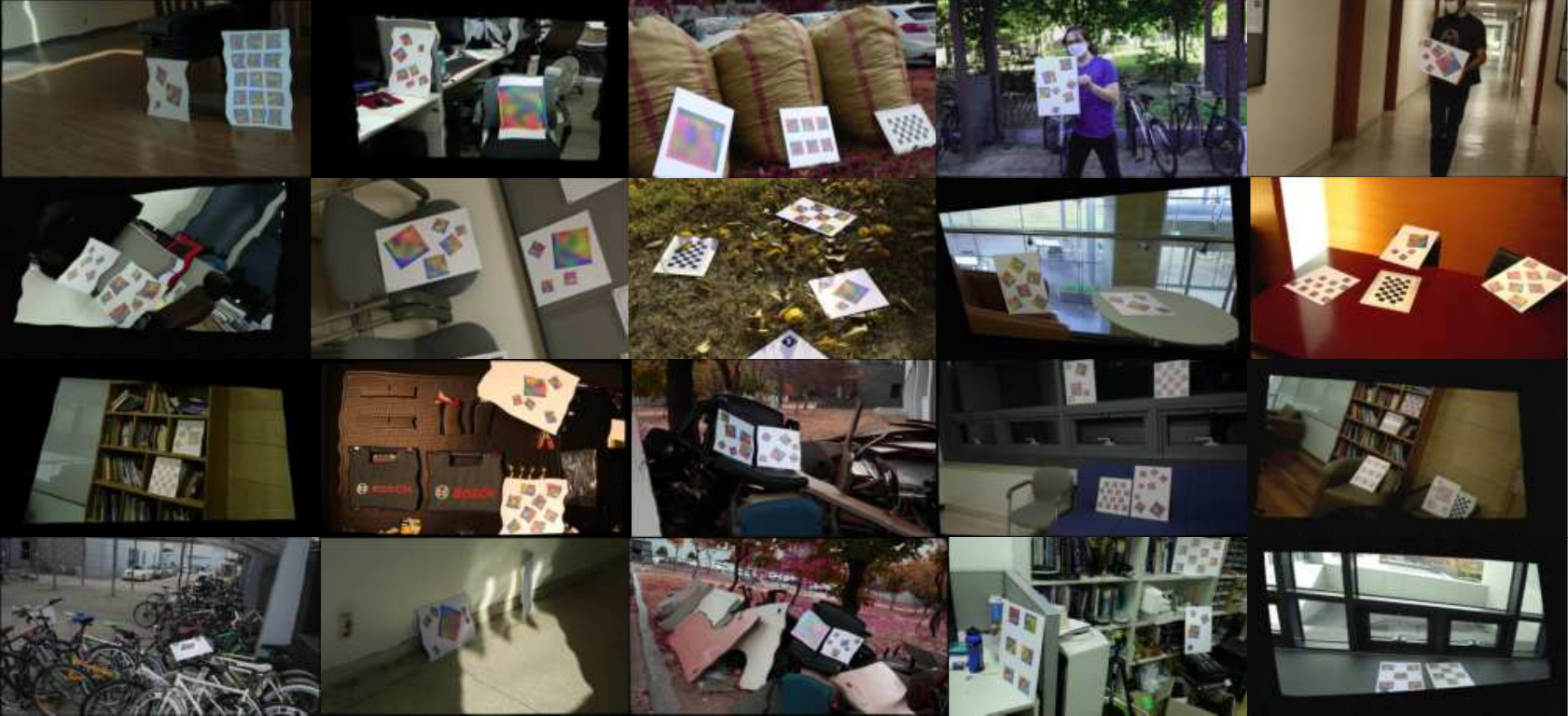}%
   \caption[]{\label{fig:training-dataset}%
	Examples of our training dataset with imaging artifact augmentation generated at runtime.}%
	\vspace{-2mm}%
\end{figure}

\paragraph{Perspective and lens distortion}
To cover \NEWB{a} more variety of perspective distortions, 
we applied random homography to the image with bilinear sampling. 
We \NEWB{also} simulate radial distortion following the Brown–Conrady model~\cite{fryer1986lens} with three random coefficients and random focal length and distortion center. 
In particular, we focus on barrel distortion (negative coefficients) as the popular wide field of view lenses are usually prone to it.

\paragraph{Color and brightness changes:}
Since our method uses color features, it is important to have \NEWB{a} realistic variety of colors and brightness of \NEWB{the} environment.
Environment illumination conditions, such as specular reflections, surface roughness, shadows, and scarce light exposure, affect the measured irradiance. 
We cover specular reflection with different roughness values, and shadows in the rendering section. 
For enhanced robustness of the detector under ill-conditioned illumination, we apply random brightness re-scale to the whole images. 
In addition, we introduce non-linear transformations through random gamma correction and also shift hues for better generalization to different lighting environments.

\paragraph{Defocus and motion blur, noise, and compression artifacts:}
We simulate imaging artifacts related to \NEWB{the} camera: defocus assuming a random-sized circular aperture and constant depth over the image and motion blur of variable magnitudes and directions. 
We also introduce Gaussian noise with a random standard deviation to mimic CMOS sensor noise with different ISO parameters. 
We finally reproduce compression artifacts through differentiable JPEG approximation~\cite{zhu2018hidden}.

After augmentations are applied, we apply gamma correction to the augmented colors convert back to sRGB values, and clamp them to the valid range for the detection network.
During training, we create random binary messages of given bits to generate fiducial markers. 
Until finishing 12 epochs, we create about 
\NEW{561,912 unique} images with all augmentations.
See Figure~\ref{fig:training-dataset} for examples of our training dataset.

\subsection{Learnable Marker Detection}
\label{sec:detection}
\paragraph{Design insight}
Our ultimate goal for marker detection is to detect \emph{multiple} markers within \NEWB{the} image 
and decode binary messages of each marker \emph{in real time}. 
To develop a novel real-time marker detector, we are inspired by the \NEWB{two-stage} Faster-RCNN approach \cite{ren2015faster}
that employs backbone features exclusively for \NEWB{both} the region proposal and classification tasks. 
\NEWB{%
In addition, we need our marker detector to be scale-invariant.
Therefore, we design two main RoI heads that are shared across the different scales of the backbone features:
(1) corner detection and
(2) message decoding heads.
To utilize backbone features efficiently, we only spatially transform them by 
inverse warping through two-stages of resampling,
and avoid costly convolution operation.
This makes our decoding network invariant to the non-linear deformation transformation,
while keeping the network depth shallow.}
\IGNORE{
Therefore, we design a novel marker detector that consists of two main RoI heads: (1) corner detection 
and (2) decoding messages by explicitly providing invariance with awareness of non-linear deformation transformation.
For efficiency, we design to fully utilize backbone features and inverse the warping through resampling two times.
We also avoid costly convolutional structures, and keep the RoI-head depths shallow. 
}

\paragraph{Network architecture}
We exclusively make use of features from the 
backbone network of object detection.
First, features are used by \NEWB{the} region proposal network (RPN) for obtaining the regions of interest (RoIs) and its initial bounding box proposal.
Second, using the bounding box proposals, features are resampled by the RoI align \cite{he2017mask} operation.
Lastly, the pooled features are directly used for (1) predicting corners by estimating affine transformation matrices and (2) decoding messages via deformation-aware sampling.
Our network architecture is shown in the rightmost column of Figure~\ref{fig:model_overview}.

We include \NEWB{Feature Pyramid Network (}FPN\NEWB{)} in our network to aid possible scale differences 
further\NEWB{~\cite{lin2017feature}}. 
With \NEWB{the} consideration of real-time performance of detection, we choose VovNet \cite{lee2019energy} as a backbone network.
\IGNORE{
RoI align provides sampled high-level features from \NEWB{the} backbone network.
We share the same feature space across the marker detector. 
Two different RoI heads infer separate outputs as (1) corners and (2) messages. 
}

\begin{figure}[tp]
\centering \footnotesize
    \includegraphics[width=0.98\linewidth]{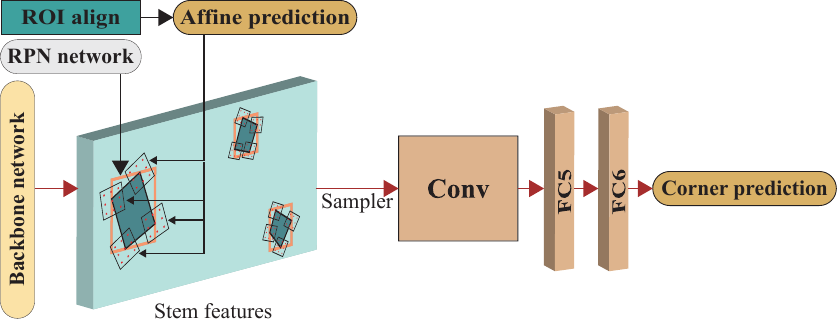}
\caption{\label{subfig:corner_head} 
Corner head. This diagram depicts the detailed architecture of our corner detection network shown in Figure~\ref{fig:model_overview}. Four affine matrices predict to sample from low-level features of the corner regions. Then the sampled regions are processed to predict relative corner coordinates in affine frame. Finally, affine matrices are used again to calculate corners in image coordinates.}
\end{figure}

\paragraph{Corner head}
Corners of markers are the critical information for various applications of fiducial markers, enabling object tracking and camera pose estimation.
However, corners are a type of information that does not include any high-level semantic meaning, and thus when we use high-level features for training the corner detector, it could not perform very well.
Accordingly, we decided to use the low-level stem features from the backbone network directly, rather than high-level FPN features. 
\NEWB{See Figure~\ref{subfig:corner_head} for a detailed architecture of our corner detector.}
For robust detection of corners, we estimate affine transformations of sampling window for each corner with respect to the center origin of the normalized RoI domain. We then sample the stem features through the transformation and then apply a convolutional and two fully connected layers to predict the corner location.  
We use the predicted corners to calculate corner loss  which is formulated as
 \begin{align}
  \label{eq:corner_loss}
     \mathcal{L}_{\text{corner}} = \frac{1}{8N_\text{total}}
     \lVert x_\text{gt}-x_\text{predict} \rVert_1,
 \end{align}
 where $N_\text{total}$ is the total number of regions, 
 $x_\text{gt}$ and $x_\text{predict}$ present pixel coordinates of ground-truth and prediction, respectively,
  and the loss computed for only foreground regions.

 \paragraph{Decoding head}
The spatial transformer network \cite{jaderberg2015spatial} has been used broadly for object classification in order to enhance classification performance by inferring optimal inverse transformation. 
However, it lacks explicit knowledge of geometric transformations.
In contrast, we actively leverage the explicit knowledge of geometric transformation applied to markers to estimate inverse transformation of deformation.
 
To this end, we first calculate uniform sampling locations in \NEWB{the} marker domain. Every time a transformation is applied, we recalculate these locations. 
We apply homography transformation for rendering, then in the augmentation stage, we apply affine augmentation, lens distortion\NEWB{,} and thin-plate spline warping consecutively and recalculate sampling locations. 
We assume that if the network can predict \NEWB{these} locations correctly, it understands the forward warping applied, and can inverse them to normalize markers back to generated marker domain. 
The sampling resolution is set to 9$\times$9, for the 12$\times$12 pooled RoI region. 
In addition, we set a sampling loss from calculated ground truth feature sample locations for decoding, which can be formulated as
 \begin{align}
 \label{eq:sampling_loss}
 \mathcal{L}_{\text{sample}} = \frac{1}{2N_\text{total}N_\text{sample}}
 \lVert x_\text{gt}-x_\text{predict} \rVert_1,
 \end{align}
 where $N_\text{sample}$ is the total number of sample locations (9$\times$9).

As a next step, we resample from the pooled RoI features using predicted sampling locations with bilinear sampling. 
Then we apply fully connected layers to the geometrically normalized marker features. 
For our comparison study of traditional markers, \NEWB{the} cross-entropy loss is used to predict the marker class. 
For our learned markers, we have decoding loss, formulated as following:
 \begin{align}
 \label{eq:decoding_loss}
\mathcal{L}_{\text{decode}} = \frac{1}{N_\text{total}N_\text{bits}}
 \lVert m_\text{encoded}-m_\text{decoded} \rVert_2^2,
 \end{align}
 where $N_\text{bits}$ is the total number of bits in a binary message,
 $m_\text{encoded}$ and $m_\text{decoded}$ present the encoded binary message of ground-truth and the decoded message, respectively. 
We also predict objectness for each RoI. We use binary cross entropy as an objectness loss $\mathcal{L}_\text{obj}$, which later used for non-maximal suppression.

Our final loss term is formulated as follows:
\begin{align}
 \label{eq:final_loss_fct}
 \mathcal{L}_\text{total} = & (\mathcal{L}_{\text{rpn}_\text{class}} + \mathcal{L}_{\text{rpn}_\text{loc}}) + \mathcal{L}_\text{sample} +\\ \nonumber
 & 0.1\mathcal{L}_\text{corner} + (0.5\mathcal{L}_\text{obj} + 10.0\mathcal{L}_\text{decode}),
\end{align}
where $\mathcal{L}_{\text{rpn}}$ loss comes from region proposal network's objectness and bounding box prediction, which is unchanged from the original detection network \cite{wu2019detectron2}.

\paragraph{Marker identification}
Once we train the end-to-end system for encoding/decoding messages, we are ready to identify markers using our marker detector network. 
From the decoder network, we receive a binary message, a 36-bit dimensional feature vector of binary codes, 
and check the identification by comparing the binary codes of each message with \NEWB{the} dictionary. 
If the confidence level of matching is lower than 80\%, we reject the identification. 
The impact of this threshold on accuracy is evaluated in Section~\ref{sec:results}.

\subsection{Implementation Details}
\label{sec:implementation_details}
We implement our method with PyTorch on the base architecture of  Detectron2~\cite{wu2019detectron2}. 
We train our models on eight NVIDIA Titan RTX GPU with Intel Xeon Silver 4214 CPU.

We employ a stochastic gradient descent with momentum and gradient clipping. 
Our model is trained with 35,000 steps ($\sim$12 epochs) with learning rate decayed to $1/10$ after 20,000 and 30,000 steps. 
We do not change the default \NEW{training} parameter settings of Detectron2.
Please refer to supplementary material for more details on network architecture.

\begin{figure}[pb]
 \centering
    \includegraphics[width=.96\linewidth]{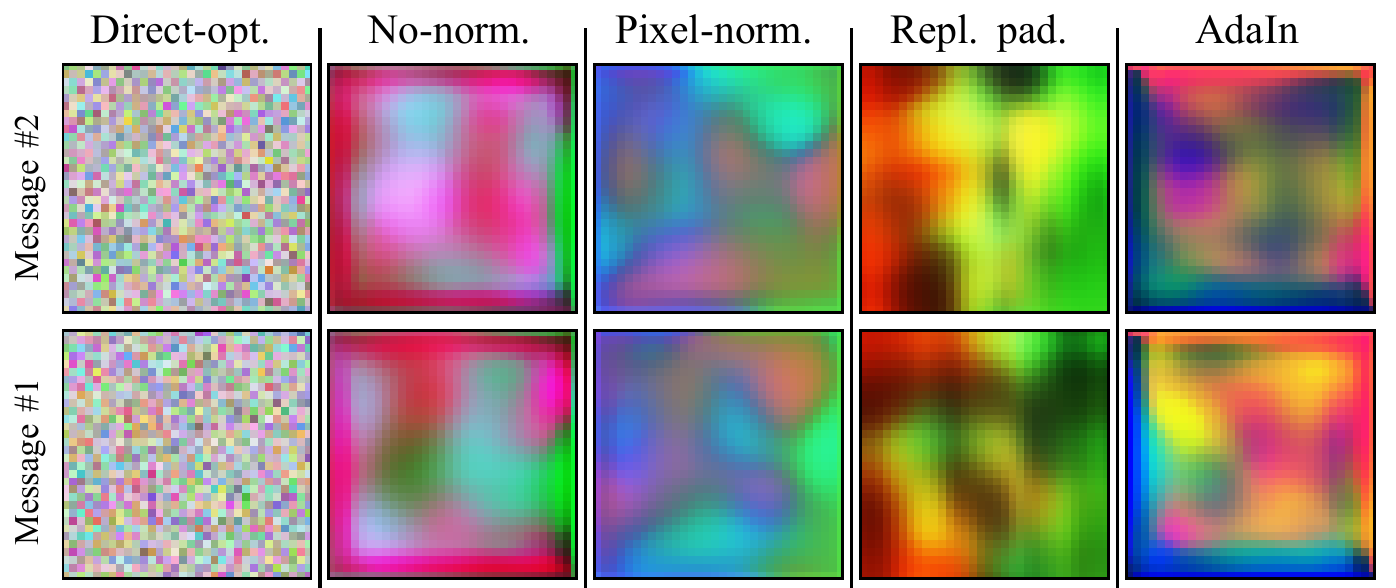}
 \caption{\label{fig:marker_qualitative} 
 Qualitative comparison using different types of optimization and normalization in our marker generator network.
 }
 \vspace{-1mm}
\end{figure}
\begin{table}[pb]
	\caption{\label{tab:generator} 
	Quantitative comparison of the generator network options in terms of initialization and normalization. The red numbers indicate the highest performance, and the blue numbers represent the second\NEWB{-}best performance.}
    \footnotesize
    \begin{center}
    \begin{tabular}{l|c|c|c|c} \hline
    \multirow{2}{*}{Method} & \multicolumn{3}{c|}{Marker Detection} & \multirow{2}{*}{Corner RMSE} \\
     & AP & {\setlength{\tabcolsep}{0pt} \begin{tabular}{c} Decoding \\ accuracy \end{tabular}} & FP Rate & \\
		\thickhline
		Direct optimization          & 75.75 &   -   & 0.0088 & 0.6699 \\
		No-norm.~generator   & 94.25 & 96.12 & \red{0.0008} & 0.3808 \\
		Pixel-norm.~generator         & \blue{94.35} & \blue{98.05} & 0.0018 & 0.3965 \\
		AdaIn generator repl. padding & 94.44 & 95.65 & \red{0.0008} & \blue{0.3653} \\ 
		AdaIn generator w/ zero pad.             & \red{95.10} & \red{98.18} & \blue{0.0014} & \red{0.3643} \\
		\hline
	\end{tabular}
    \end{center}
	\vspace{-2mm}
\end{table}

\begin{table*}[pth]
    \caption{\label{tab:rendered_main} 
	 To evaluate the performance of our image simulator independently from other parts of our system, 
	 we render Aruco markers as shown in Figure~\ref{fig:rendering_abl_dataset} and train the same decoder network. 
	 This table shows $AP$ metric evaluation of each option of our image simulator. 
	 Our detector model is trained on different rendering methods, indicated as columns. 
	 Simulated and real-world test datasets are used to evaluate which indicated as two main rows. 
	 Each sub-row and column indicate the augmentations used to train/test the models. 
	 As the simulated and real test datasets yields high accordance, simulated images are used to evaluate \NEWB{the} architectural decisions of the network. 
	 Real-world testing rows indicate that our full simulation pipeline with training augmentations yield the best results across all testing categories. 
	 Including color and geometric augmentations helps the generalization of the models across all datasets.
	 Note that the overall scores on fully augmented test datasets is still low as imaging artifacts impose challenges
	 to detection and due to the absence of our learnable markers.}
    \centering \footnotesize
    \setlength{\tabcolsep}{2.3pt}
	\begin{tabular}{ll|c|c|cccc|cccc|cccc} \hline
		\multicolumn{2}{l|}{\multirow{2}{*}{\diagbox{Test}{Train}}} &
		    \multicolumn{1}{c|}{\underline{Baseline}} &
		    \multicolumn{1}{c|}{\underline{Superimp.}} &
		    \multicolumn{4}{c|}{\underline{Placement w/o shading}} &
		    \multicolumn{4}{c|}{\underline{Diffuse shading only}} &
		    \multicolumn{4}{c}{\underline{Diffuse shading + spec.}}\\
		 & & Aruco & COCO & No & Color & Geo & Full & No & Color & Geo & Full & No & Color & Geo & Full \\
		\thickhline
		\multirow{4}{*}{\rotatebox[origin=c]{90}{\underline{Simulated}}}
		& No &
		    70.41 & 29.76 &
		    02.47 & 57.40 & 04.05 & 60.72 & 
		    10.32 & 75.24 & 19.74 & 76.95 &
		    79.13 & \blue{91.07} & 87.47 & \red{92.80} \\
		& Color &
		    37.55 & 16.54 &
		    00.40 & 32.97 & 00.40 & 34.56 & 
		    05.72 & 68.00 & 11.27 & \blue{71.15} &
		    45.61 & 70.06 & 43.37 & \red{72.19} \\
		& Geo. &
		    35.82 & 22.98 &
		    01.16 & 48.02 & 02.60 & 53.79 & 
		    05.11 & 64.24 & 14.75 & 69.80 &
		    54.51 & \blue{77.65} & 70.50 & \red{82.44} \\
		& Full &
		    22.89 & 13.50 &
		    00.28 & 27.70 & 00.22 & 30.23 & 
		    03.11 & 54.25 & 08.08 & \blue{59.09} &
		    33.84 & 56.19 & 34.95 & \red{59.78} \\
		\thickhline
		\multirow{4}{*}{\rotatebox[origin=c]{90}{\underline{Real}}}
		& No &
		    67.60 & 35.59 &
		    02.89 & 51.42 & 04.94 & 52.30 & 
		    27.18 & 76.92 & 55.54 & 80.52 &
		    57.96 & \blue{85.24} & 68.72 & \red{87.66} \\
		& Color &
		    36.20 & 17.39 &
		    00.02 & 25.88 & 00.07 & 26.29 & 
		    11.97 & 55.10 & 23.47 & 58.35 &
		    27.04 & \blue{60.77} & 33.40 & \red{62.45} \\
		& Geo. &
		    34.60 & 26.62 &
		    01.06 & 41.02 & 03.19 & 44.88 & 
		    15.21 & 64.31 & 42.30 & 70.94 &
		    41.42 & \blue{71.51} & 54.70 & \red{76.73} \\
		& Full &
		    22.29 & 14.43 &
		    00.00 & 21.63 & 00.00 & 22.91 & 
		    07.95 & 44.17 & 18.69 & 48.49 &
		    27.47 & \blue{48.87} & 27.14 & \red{51.92} \\ \hline
	\end{tabular} 
	 \vspace{-2mm}
\end{table*}

\section{Ablation Study}
\label{sec:ablations}
For comprehensive ablation experiments, we choose our synthetic dataset including the ground truth values of corners and messages.

\paragraph{Evaluation metric}
To evaluate the accuracy of marker detection results, we employ a popular metric for object detection, 
average precision (AP)\footnote{Average precision (AP)  computes the average precision value (about how accurate the predictions are)
for recall value (how good the detector \NEWB{is} finding \NEWB{the positive examples})
in the percentage scale. AP is averaged over all markers in our experiments.} 
instead of object proposal proxy metrics, following other detection works~\cite{ren2015faster}. \NEW{We make use of Detectron2 implementation to calculate metrics.}

\subsection{Marker Generation}
\label{sec:generation}
We train our marker generation network with different options in the marker generation architecture.
Figure~\ref{fig:marker_qualitative} compares the visual markers qualitatively, 
and Table~\ref{tab:generator} evaluates AP scores, message decoding accuracy, false positive (FP) rate, and corner errors.
As shown in the leftmost part of Figure~\ref{fig:model_overview}, 
our marker generator consists of five main blocks: FC block, three sub-generation blocks, and color conversion block. 
Each sub-generation block consists of upsampling, convolution, and normalization.
We evaluate the impact of different normalization options: no normalization, pixel-wise feature normalization, 
channel-wise AdaIn normalization with replication padding, and AdaIn normalization with zero padding. 
Also, we compare them with direct optimization, which is to calculate gradients of randomly initialized 
weights for optimization of marker patterns.

\NEWB{Quantitatively},
we \NEWB{find that} 
the AdaIn normalization with zero padding
outperforms all other options in terms of AP score and decoding accuracy in marker detection. 
It also outperforms others in terms of corner detection accuracy.
We\NEWB{,} therefore\NEWB{,} 
use the AdaIn normalization with zero padding \NEWB{as} 
our method.
\NEWB{%
Qualitatively, we observe that the learned markers produce common features, such as 
blue patterns at the left-bottom corner, and thus preserving appearance similarity. 
We found that the rich variety of marker appearances on our selected generation method
 supports our design insights described in Section~\ref{subsec:marker_gen}.}

\subsection{Image Simulation}
\label{sec:image-simulation}
\begin{figure}
    \centering  \footnotesize
    \begin{subfigure}{.15\textwidth}
    \includegraphics[trim={8cm, 4cm, 8cm, 1.5cm}, clip, width=.97\linewidth]{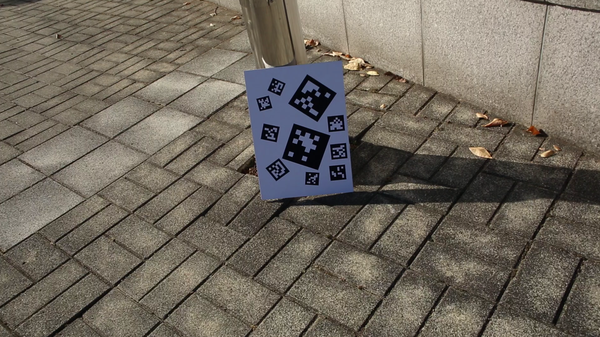}
    \caption{\label{fig:rendering_abl_raw} Captured}
    \end{subfigure}
    \begin{subfigure}{.15\textwidth}
    \includegraphics[trim={8cm, 4cm, 8cm, 1.5cm}, clip, width=.97\linewidth]{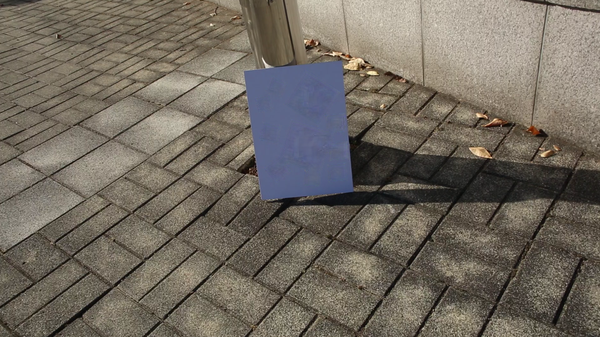}
    \caption{\label{fig:rendering_abl_inp} Inpainted}
    \end{subfigure}
    \begin{subfigure}{.15\textwidth}
    \includegraphics[trim={8cm, 4cm, 8cm, 1.5cm}, clip, width=.97\linewidth]{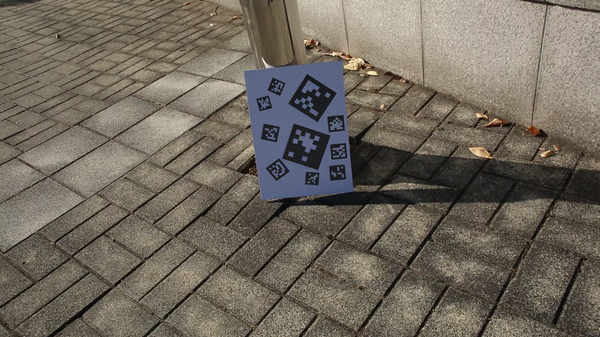}
    \caption{\label{fig:rendering_abl_rendered} Simulated}
    \end{subfigure} 
    \caption{\label{fig:rendering_abl_dataset} Rendering pipeline for creating Aruco images for image simulation ablation (Table~\ref{tab:rendered_main}). We first capture boards with random Aruco markers~(\subref{fig:rendering_abl_raw}), we then simulate the same markers at the same location through our inpainting~(\subref{fig:rendering_abl_inp}) and rendering~(\subref{fig:rendering_abl_rendered}) approach.
    }
\end{figure}

To evaluate our image simulator, we train our detection network with different rendering configurations.
Note that we employ Aruco markers to exclude the impact of the marker generator network.
We then test our models on both the captured boards with random Aruco markers and a rendered image dataset.
The rendered images are simulated by our renderer with the same placement configuration as the real-world dataset (see Figure~\ref{fig:rendering_abl_rendered}), creating a simulated replica of the real-world test dataset.
The model trained on our diffuse and specular rendering combined with geometric and color augmentations shows the best results for both synthetic and real-world evaluation datasets (Table~\ref{tab:rendered_main}).
In addition,  we apply augmentation to both test datasets, making the detection task significantly more challenging. 
Table~\ref{tab:rendered_main} reveals that training with augmentation successfully improves \NEWB{the} robustness 
of the detection networks, for all rendering methods.
Evaluation results indicate that the simulated and the rendered test datasets yield
consistent gap with mean difference of 2.455 and standard deviation of 10.066. 
This shows that rendering can be used for further ablation studies.

\subsection{Marker Detection}
\label{sec:marker-detection}
\paragraph{Backbone networks}
We experiment on different backbone networks, to select the most suitable architecture for our real-time marker detection system. 
We select networks with feature pyramid structure to aid better for scale differences, 
mainly ResNet \cite{he2016deep}, and VoVNet \cite{lee2019energy} with different depths. 
Prior to our experiments, we expected deeper networks to perform better at recognition task\NEWB{s}. 
On the contrary, we observe that shallow VoVNet19 DW-FPNLite provide\NEWB{s} better performance than the others. 
See Table~\ref{tab:backbone} for comparison.

\begin{table}[tp]
	\caption{\label{tab:backbone}
	We evaluate the impact of different backbone network architectures for marker detection performance. We test two different levels of ResNet with FPN, and VoVNet. 
	Despite the shallower depth of the network, we found that VoVNet19 DW-FPNLite 
	\cite{lee2019energy} is the best.}
    \centering \footnotesize
    \setlength{\tabcolsep}{3.5pt}
    \begin{tabular}{l|c|ccc|c|c|c} \hline
	    Method & $AP$ & $AP_{S}$ & $AP_{M}$ & $AP_{L}$ & 
	    	{\setlength{\tabcolsep}{0pt}
	    		\begin{tabular}{c} Decoding \\ accuracy \end{tabular}}
	    	 & FP-Rate & FPS \\
		\thickhline
		ResNet50 FPN       & 92.40 & 88.83 & \red{99.51} & 100 & 97.91 & \red{0.0014} & \blue{17.0} \\
		ResNet100 FPN       & 90.45 & 86.00 & 99.25 & 100 & 97.68 & \blue{0.0018} & 12.5 \\
		VoVNet39 FPN        & \blue{94.38} & \blue{92.22} & 98.85 & 100 & \red{98.40} & 0.0022 & 16.5 \\
		VoVNet19 DW-FPNLite & \red{95.10} & \red{93.06} & \blue{99.43} & 100 & \blue{98.18} & \red{0.0014} & \red{29.0} \\ \hline
	\end{tabular}
	\vspace{-3mm}
\end{table}

\begin{figure}
	\centering  \footnotesize
	\begin{subfigure}{.205\textwidth}
		\includegraphics[width=.98\linewidth]{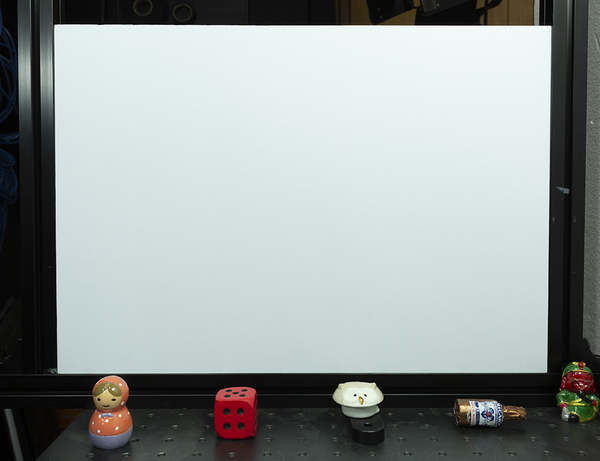}
		\vspace{-2mm}
		\caption{\label{fig:messaging_raw} Base real image}
	\end{subfigure}
	\begin{subfigure}{.205\textwidth}
		\includegraphics[width=.98\linewidth]{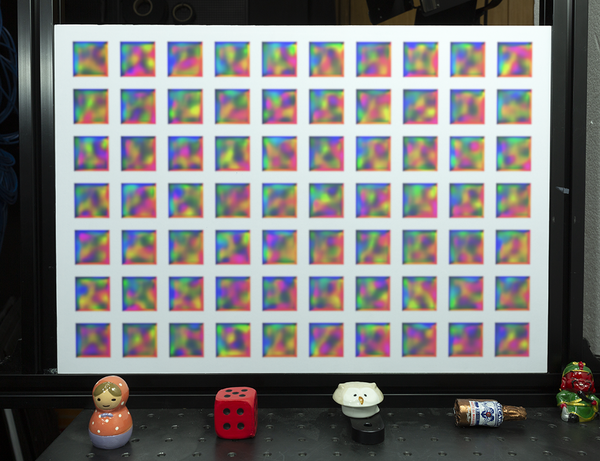}
		\vspace{-2mm}
		\caption{\label{fig:messaging_rendered} Rendered with our markers}
	\end{subfigure} 
	\vspace{-2mm}
	\caption{
		\label{fig:messaging_image} 
		\NEW{Experimental setup for testing the messaging capability of our method. We capture an empty white board (a) to render our fiducial markers (b) through iterations of encoding/decoding of messages.}
			\vspace{0mm}
		}
\end{figure}

\paragraph{Message embedding capacity}
\NEW{%
	To evaluate the message embedding capability and impact of the number of bits on decoding accuracy, 
	we train models with three different message lengths of 18, 36, and 64~bits.
	We place different markers to the image using our rendering pipeline as shown in Figure~\ref{fig:messaging_raw}.
	We place 70 markers at each iteration (see Figure~\ref{fig:messaging_rendered}).
	We test all possible markers for the 18-bit model and test $\sim$1 billion (=$2^{30}$) random markers for 36- and 64-bit models. This number of samples is limited by the computing resource in our setup (Section~\ref{sec:implementation_details}).
	These experiments took 12 days.
	For fair validation of the 36- and 64-bit models, we employ stratified sampling of bit messages to cover all marker feature space with $\sim$4.2 million markers sampled for 256 subgroups.
	Statistical results provided in Table~\ref{tab:messaging} validate that the generated markers are encoded and decoded successfully with high accuracy within the bit range of messages.}

\begin{table}
	\caption{\label{tab:messaging} 
		\NEW{Decoding accuracy results over markers  sampled from three different models of 16-, 36-, and 64-bit models.
			Percentages of 0-bit error (all bits in each message are correctly predicted) and 1-bit error are also presented.}}
    \centering \footnotesize
    \setlength{\tabcolsep}{3.5pt}
	\begin{tabular}{l|c|c|c|c} \hline
	Model & Mean decoding accuracy & Standard deviation & 0-bit error & 1-bit error \\
	\thickhline
	16 bits & 99.998 & 0.1143 & 99.97\% & 0.03\% \\
	36 bits & 99.921 & 0.7605 & 98.51\% & 00.75\% \\
	64 bits & 99.558 & 1.0787 & 80.39\% & 14.00\% \\
	\hline
	\end{tabular}
	\vspace{-2mm}
\end{table}		

\paragraph{RoI heads}
\NEW{We investigate a variety of RoI heads on detection and decoding performance with our learned markers. 
We found that na{\"i}ve convolutional decoding head without any learned spatial transformation fails to converge, 
and thus we instead employ an affine-based spatial transformer network \cite{jaderberg2015spatial} as a baseline.}
Next, we develop our special decoding head that uses supervision signals to inverse the transformations. 
In addition, to provide better corner accuracy, our corner head learns to adaptively sample the corner regions, and predict the corner locations. 
\NEWB{%
Table~\ref{tab:roi_heads} shows that our guided transformer improves the decoding accuracy significantly
by only resampling the backbone features without expensive operation. 
It keeps our network efficient, and satisfies our design insights described in Section~\ref{sec:detection}.
In addition, the affine-based sampled low-level backbone features, utilized by the corner head, improve
sub-pixel corner prediction accuracy with a significant margin.}
\IGNORE{
Our ablation results show that our new architectures improve\OLD{s} our detection capability
and sub-pixel corner prediction accuracy by a significant margin.
Table~\ref{tab:roi_heads} \NEW{compares the results}.
}

\begin{table}[htpb]
	\caption{\label{tab:roi_heads} 
	RoI head architecture validation. We compare the marker detection accuracy and corner prediction accuracy. Our guided transformer and affine-based RoI head outperform the baseline methods.}
	\setlength{\tabcolsep}{1.8pt}
    \footnotesize
    \begin{center}
    \begin{tabular}{l|c|c|c|c}\hline
	    Method & $AP$  & 
	    	{\setlength{\tabcolsep}{0pt}
	    	\begin{tabular}{c} Decoding \\ accuracy \end{tabular}} 
    	& FP-Rate & 
    		{\setlength{\tabcolsep}{0pt}
    		\begin{tabular}{c} Corner \\ RMSE \end{tabular}}  \\
		\thickhline
		Baseline corner head + spatial transformer  & 83.69 & 95.41 & 0.0022 & 1.0426 \\
		Baseline corner head + guided transformer     & \blue{89.80} & \blue{97.09} & \blue{0.0015} & \blue{0.7768} \\
		Affine-based corner head + guided transformer & \red{95.10} & \red{98.18} & \red{0.0014} & \red{0.3643} \\ \hline
	\end{tabular}
    \end{center}
\end{table}

\begin{figure}
	\centering  \footnotesize
	\begin{subfigure}{.155\textwidth}
		\includegraphics[trim={1cm, 3cm, 10cm, 3cm}, clip, width=.99\linewidth]{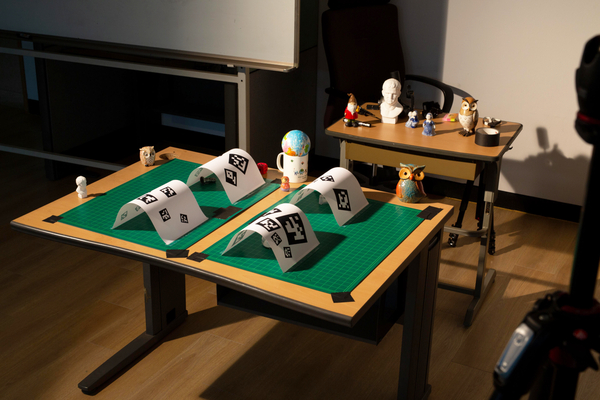}
		\vspace{-4mm}
		\caption{\label{fig:deformation_example1}\footnotesize Detection with Aruco}
	\end{subfigure}
	\begin{subfigure}{.155\textwidth}
		\includegraphics[trim={1cm, 3cm, 10cm, 3cm}, clip, width=.99\linewidth]{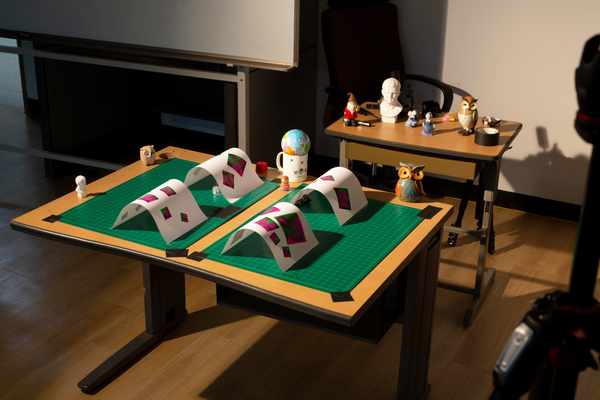}
		\vspace{-4mm}
		\caption{\label{fig:deformation_example2}\footnotesize Detection without TPS}
	\end{subfigure} 
	\begin{subfigure}{.155\textwidth}
		\includegraphics[trim={1cm, 3cm, 10cm, 3cm}, clip, width=.99\linewidth]{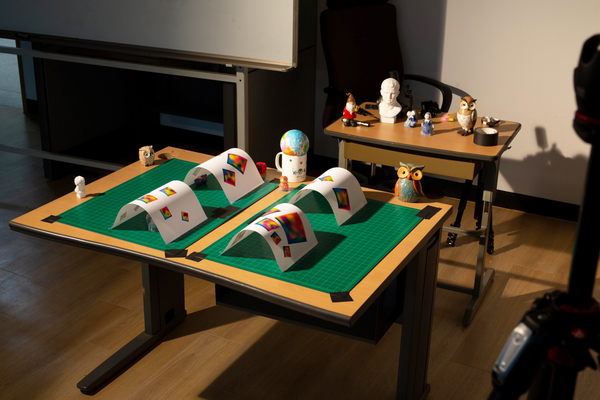}
		\vspace{-4mm}
		\caption{\label{fig:deformation_example3}\footnotesize Detection with TPS}
	\end{subfigure}
	\caption{%
		\label{fig:deformation_dataset} 
		\NEW{We build an experimental setup to evaluate the impact of our RoI head network and TPS-based augmentation for detecting markers.}}
	\vspace{2mm}%
	\includegraphics[width=.995\linewidth]{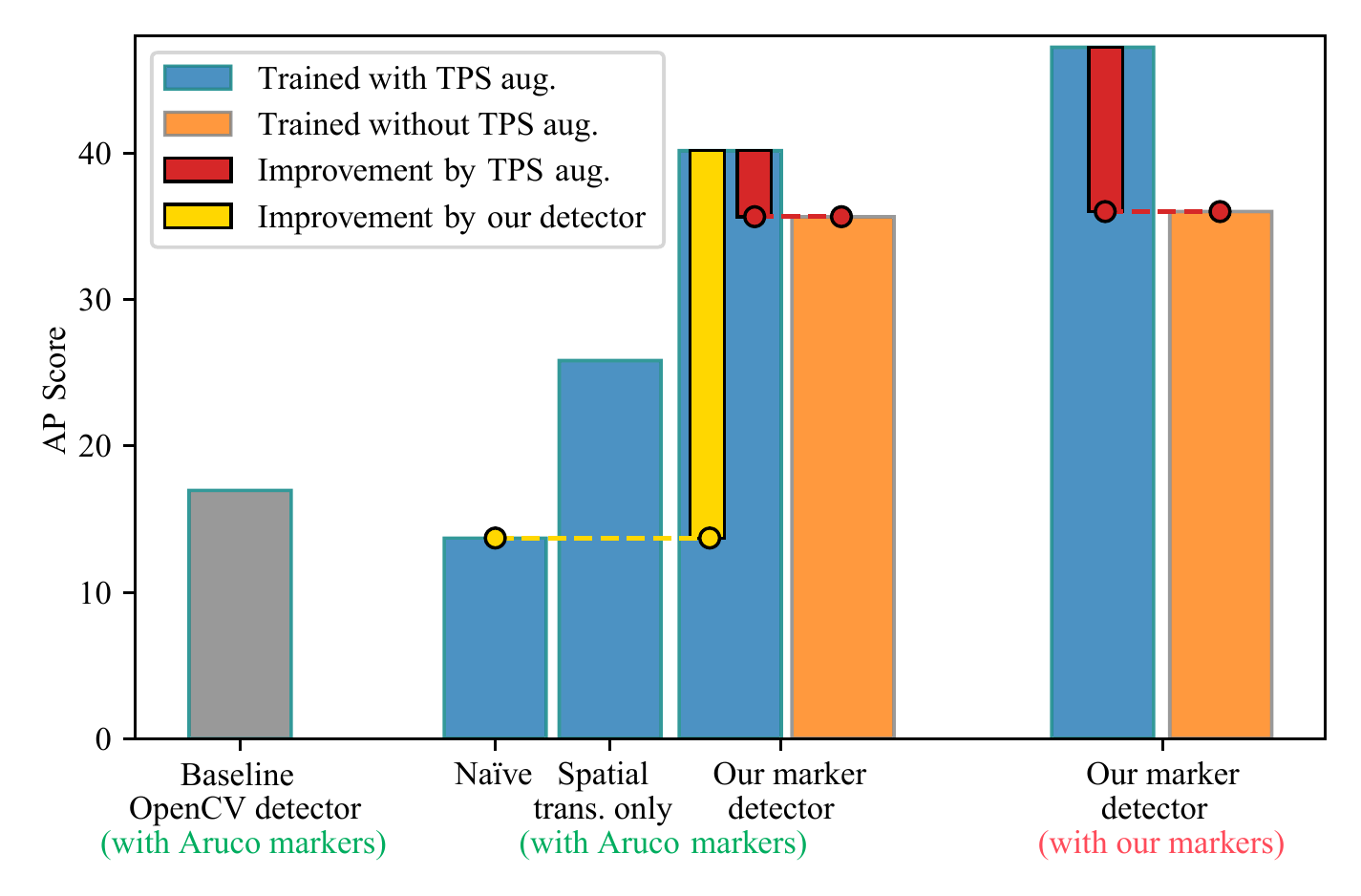}%
	\vspace{-1mm}%
	\caption{\label{fig:tps_ablation}
	\NEW{Using the experimental setup shown in Figure~\ref{fig:deformation_dataset},
	we compare the impact of our marker detection network and our training with TPS-based augmentation on detection accuracy with deformed fiducial markers. As shown in results using Aruco markers in the middle, our detection network contributes more than our TPS-based augmentation for robust detection performance of deformed markers. When our learnable markers are applied additionally to our detector method, our system outperforms all the baseline methods significantly.}}
	\vspace{-2mm}
\end{figure}

\NEW{%
\subsection{Deformation}
\label{sec:deformation-ablation}
Our marker detection network consists of two main blocks: the guided transformer network and the corner prediction network.
Our detection network is also trained with TPS-based augmentation.
In this study, we evaluate the relative impact of our detection network and the TPS-based augmentation in training using the traditional Aruco markers.  
We built an experimental setup, where Aruco and our markers trained with/without TPS augmentation are printed on paper and deformed as shown in Figure~\ref{fig:deformation_dataset}.

Figure~\ref{fig:tps_ablation} compares detection performances with different configurations, compared with the baseline OpenCV detector with Aruco markers.
In the middle, the four different bars present AP scores  using four different configurations with Aruco markers.
The first blue bar shows the AP score using only a vanilla VoVNet19 network and Aruco markers. The simple detection network does not account for any spatial deformation, resulting in a lower performance than the baseline method.
The second blue bar presents an improvement achieved by applying an existing 
affine-based spatial transformer network~\cite{jaderberg2015spatial} in the RoI head.
The third blue bar shows the performance of our detector network with Aruco markers, being trained with TPS-based augmentation.
The fourth orange bar compares the result by the same network but trained without TPS-based augmentation.
As indicated by the additional yellow and red thin bars, 
when we employ the fixed patterns of Aruco markers, the improvement by our detection network is more significant than that by our TPS-based augmentation in training. 

In addition, we compare these Aruco results with our learnable marker results. 
Unfortunately, the training process of our marker detector without any spatial 
transformation network does not converge so that we cannot compare the result. 
Instead, we compare the precision results of our detection network trained with/without TPS-based deformation. 
As shown in the rightmost part \NEWB{of} the plot, when our method is trained with our learnable markers and TPS-based augmentation, 
our marker detector outperforms both the baseline detector and the Aruco-based trained network significantly. 
}

\section{Results}
\label{sec:results}

\begin{figure}[bhtp]
 \centering
    \includegraphics[width=\linewidth]{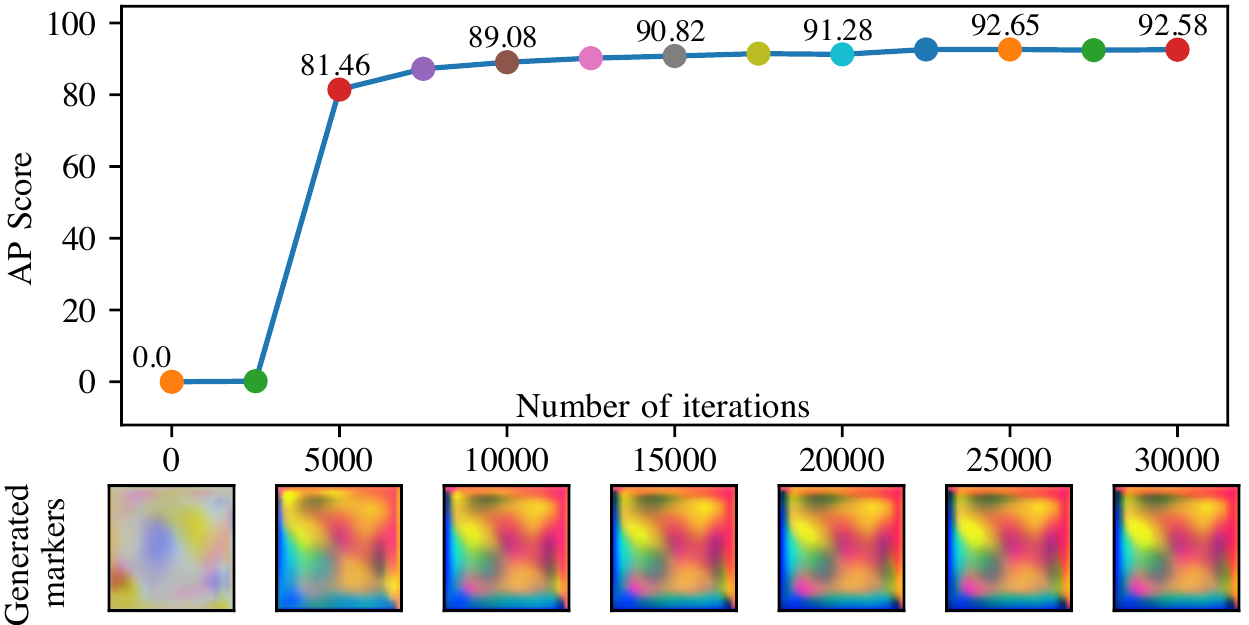}
    \caption{\label{fig:accuracy_plot} 
    Marker appearance changes over iterations. With more iterations, the accuracy increases asymptotically and the marker generator uses more saturated colors.}
 \vspace{-3mm}
\end{figure}

\paragraph{Markers' appearance over iterations}
We evaluate \NEWB{increasing} 
AP scores over every 2,500 iterations. 
For \NEW{every} 5,000 iterations, we also evaluate the visual appearance of one of the optimized visual markers.
Figure~\ref{fig:accuracy_plot} presents testing AP scores with our synthetic test dataset, and also visualizes the marker's appearances over the training iterations.
\NEWB{Starting from 5,0000 iterations, the network achieves reliable AP scores.}
The appearance of the optimized marker becomes stabilized after 25,000 to 30,000 iterations.  
With more iterations, the accuracy increases asymptotically and the marker generator uses more saturated colors and defines clearer lines at the boundaries.

\begin{figure}[htpb]
 \centering
    \includegraphics[width=.95\linewidth]{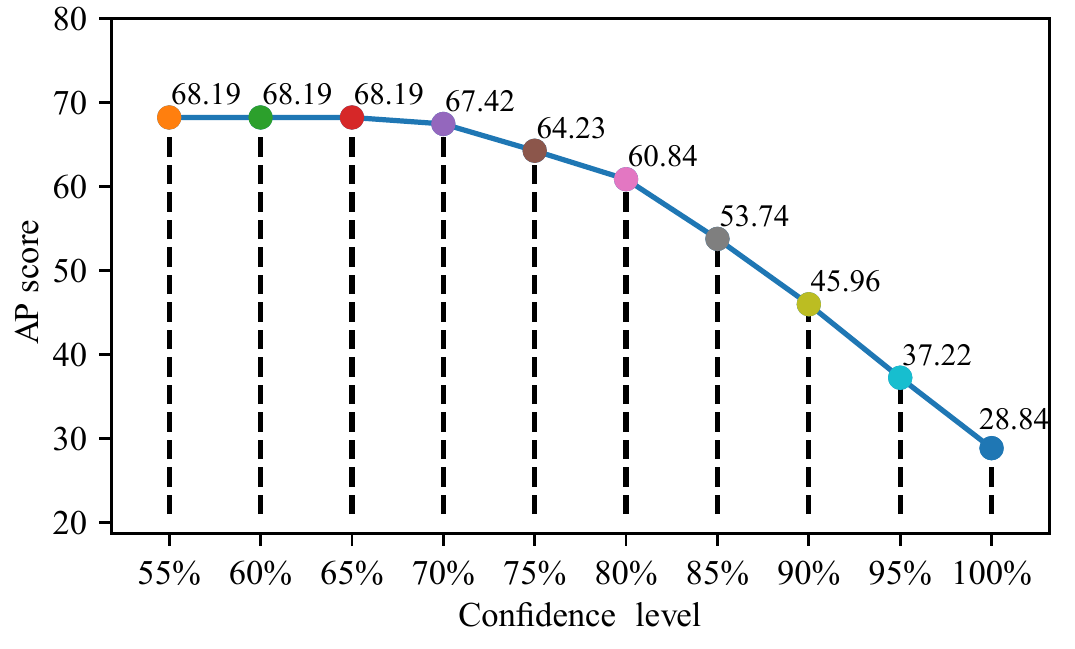}
    \caption{\label{fig:confidence-level} Impact of the confidence level on the AP score tested with the real-world test dataset. We choose 80\% as the confidence level threshold for our experiments.}
 \vspace{-1mm}
\end{figure}

\paragraph{Impact of the confidence level}
Once we estimate a binary message code from a detected marker, we apply confidence filtering to determine the marker's identification. 
In our current system, the confidence level is a hyperparameter in our detection mechanism. 
We evaluate the impact of the confidence level in terms of AP score with our real-world test dataset. 
There is a tradeoff between \NEWB{the} AP score and  confidence level. 
When we increase the level of confidence, the overall AP score decreases. When the confidence level decreases, 
it potentially increases \NEWB{the} risks of having more false positives. 
We, therefore, choose 80\% for the confidence level in our experiments. See Figure~\ref{fig:confidence-level}.

\begin{figure}
	\centering  \footnotesize
	\includegraphics[ width=.97\linewidth]{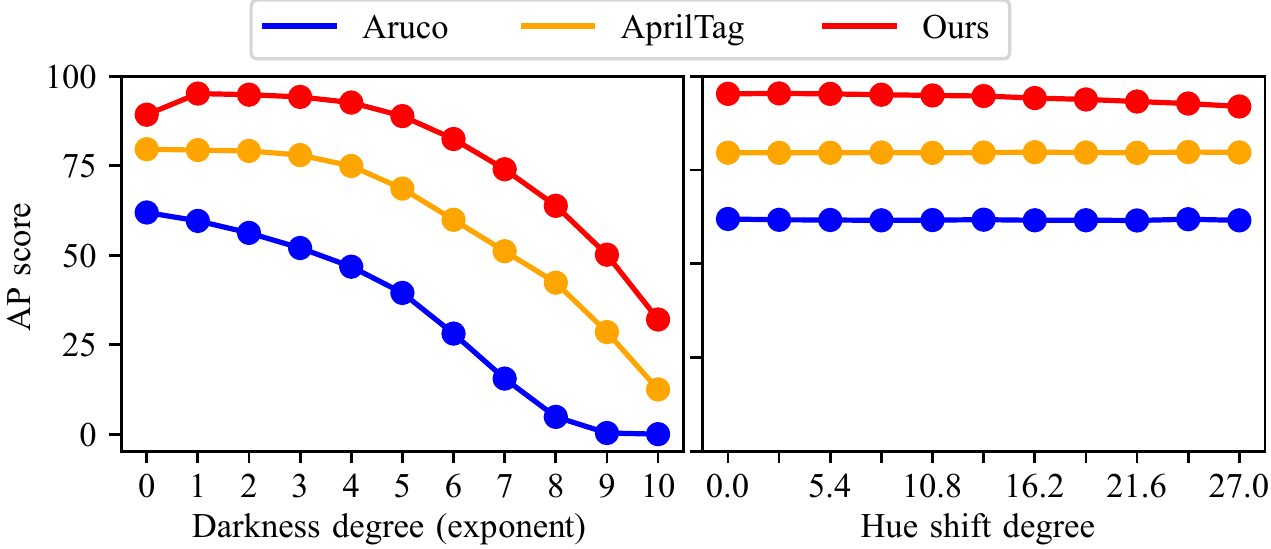}
	\caption{\label{fig:hue_bright_ap}
	\NEW{To evaluate the robustness of three methods: Aruco, AprilTag, and ours, we create brightness changes and hue shifts synthetically and compare precision performances. Brightness is scaled by $0.6^k$, with $k$ ranging from 0 to 10 in the linear intensity domain. The hue values of pixel colors are shifted by 0 to 27 degrees in HSV color space. Our method outperforms the baseline methods within the range of scene illumination conditions.}}
\end{figure}

\paragraph{\NEW{Robustness against brightness and hue changes}}
\NEW{Instead of binary black-and-white features, we create color features using our generative network, and thus its performance might be affected by brightness or hue changes of scene illumination. 
We, therefore, evaluate the performance of our detection network by synthetically varying the level of brightness and hue of images. 
We compare our method's AP scores with those of binary methods: Aruco \cite{munoz2012aruco} and AprilTag \cite{wang2016apriltag}.
For brightness changes, we scale the level of pixel intensity by $0.6^k$, with $k$ ranging from 0 to 10 in the linear intensity domain.
For hue changes, we shift the hue values of pixel colors by 0 to 27 degrees in HSV color space.
As shown in Figure~\ref{fig:hue_bright_ap}, our method still outperforms these two baseline methods of Aruco (implemented by OpenCV) and AprilTag (implemented by the authors).}

\begin{figure}[tp]
	\centering
	\begin{subfigure}{.115\textwidth}
		\includegraphics[trim={12.5cm, 1.5cm, 1.5cm, 3cm}, clip, width=.97\linewidth]{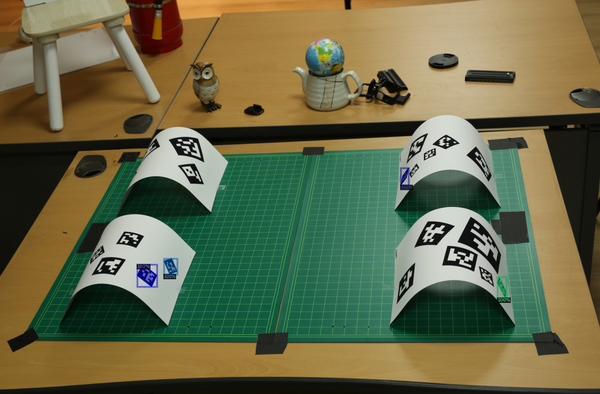}
		\vspace{-1mm}
		\caption{\label{fig:aruco_deform} Aruco}
	 \end{subfigure}
	 \begin{subfigure}{.115\textwidth}
		\includegraphics[trim={12.5cm, 1.5cm, 1.5cm, 3cm}, clip, width=.97\linewidth]{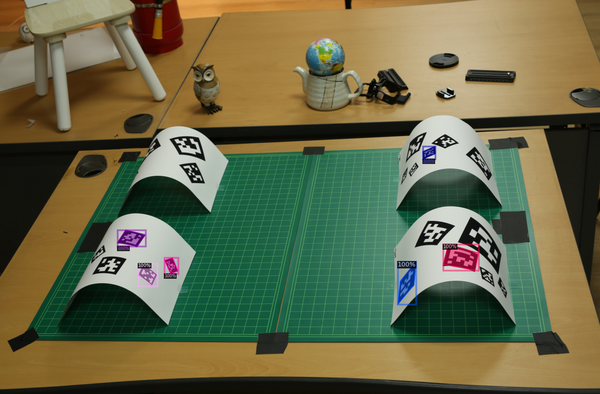}
		\vspace{-1mm}
		\caption{\label{fig:april_deform} AprilTag}
	 \end{subfigure}
	  \begin{subfigure}{.115\textwidth}
		\includegraphics[trim={12.5cm, 1.5cm, 1.5cm, 3cm}, clip, width=.97\linewidth]{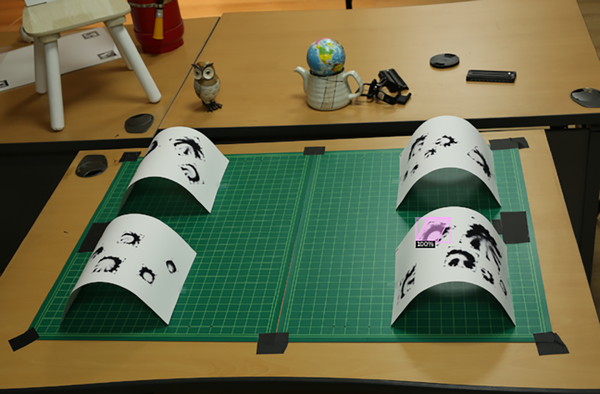}
		\vspace{-1mm}
		\caption{\label{fig:e2e_deform} E2ETag}
	 \end{subfigure}
	 \begin{subfigure}{.115\textwidth}
		\includegraphics[trim={12.5cm, 1.5cm, 1.5cm, 3cm}, clip, width=.97\linewidth]{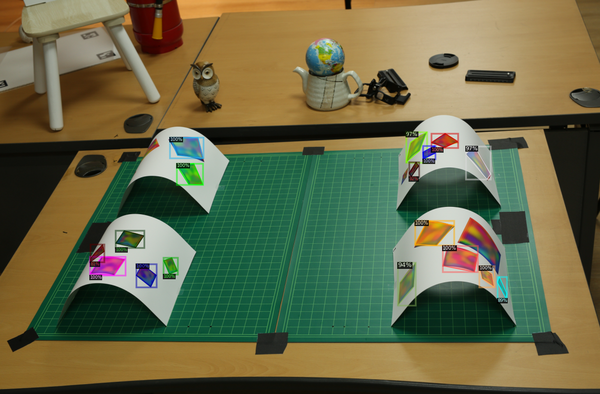}
		\vspace{-1mm}
		\caption{\label{fig:ours_deform} Ours}
	 \end{subfigure}
	\caption{\label{fig:comparison_qualitative} Setup for deformation evaluation. 
	We bend the paper with the same distances (10\,cm) for all methods to  analyze the robustness of each method to placement on non-flat surfaces. 
	}
	\vspace{-1mm}
\end{figure}

\begin{figure*}[bhtp]
	\centering \footnotesize
	\begin{subfigure}{.49\textwidth}
		\includegraphics[trim={0cm, 1cm, 0cm, 2cm}, clip, width=0.99\linewidth]{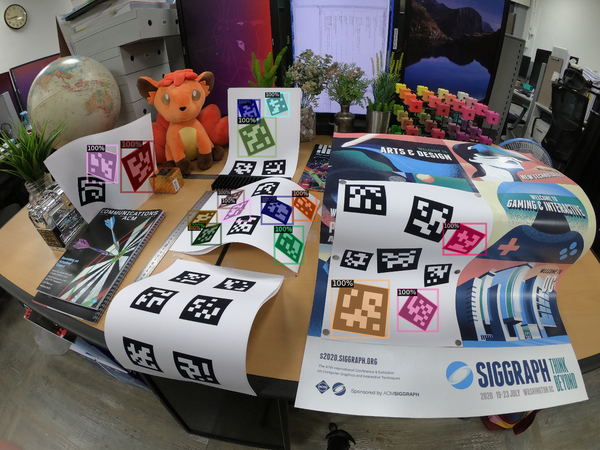}
		\vspace{-1mm}
		\caption{\label{fig:comparison2_april} AprilTag on deformed surfaces}
	\end{subfigure}
	\begin{subfigure}{.49\textwidth}
		\includegraphics[trim={0cm, 1cm, 0cm, 2cm}, clip, width=0.99\linewidth]{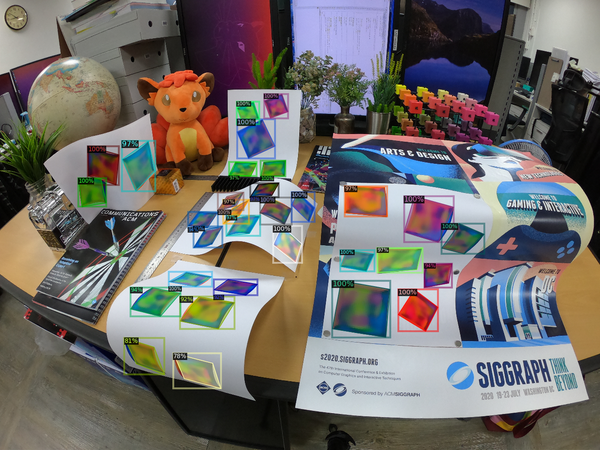}
		\vspace{-1mm}
		\caption{\label{fig:comparison2_ours} Our markers on deformed surfaces}
	\end{subfigure} 
	\vspace{-1mm}
	\caption{\label{fig:comparison2} Qualitative comparison of our deformable markers against AprilTag on deformed surfaces. While AprilTag fails on deformed markers, our method detects all the markers on deformed surfaces.}
	\vspace{-1mm}
\end{figure*}

\begin{figure}[bhtp]
	\centering
	\vspace{-2mm}
\includegraphics[width=.85\linewidth]{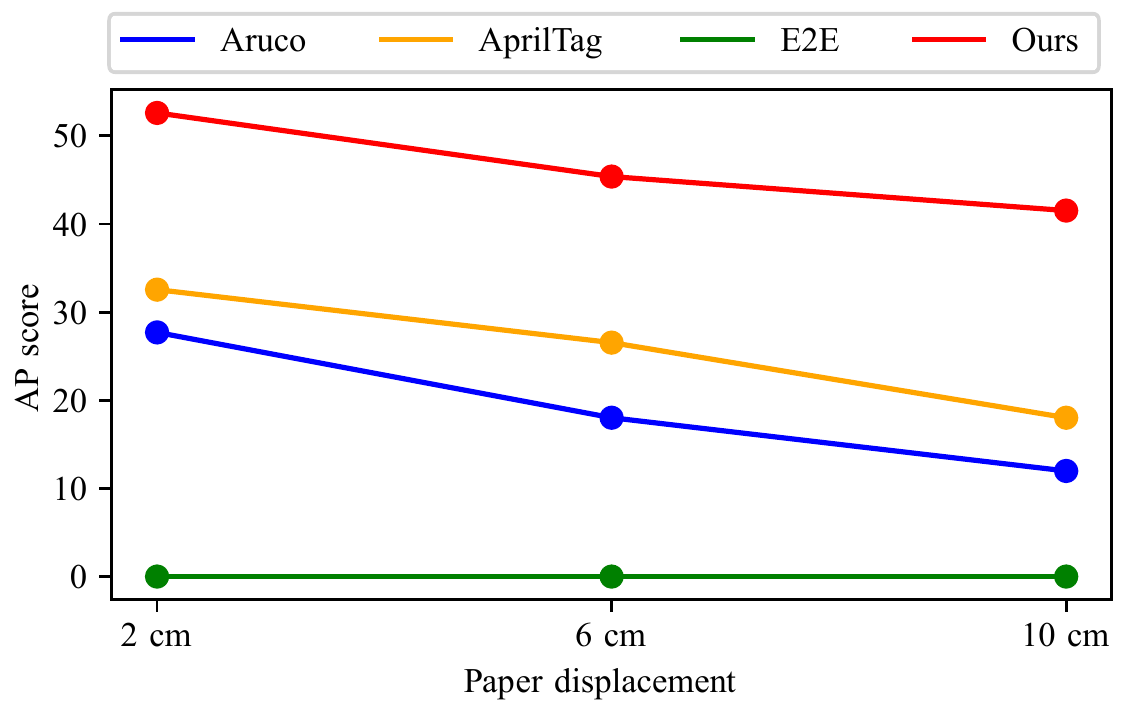}%
	\caption{\label{fig:paper-displacement} AP score against different deformations. A larger displacement yields a stronger paper curvature. Our method significantly outperforms others when the the markers are placed on non-planar surfaces.}
	\vspace{-1mm}
\end{figure}

\paragraph{\NEW{Real-world deformation experiments}}
\NEWB{To} 
evaluate the performance of these four methods on deformed surfaces, we design a controlled environment for a marker deformation experiment.
We fix one side of the paper on a flat table and move the other side to bend the sheet and create a consistent saddle shape (see Figure~\ref{fig:comparison_qualitative}).
We prepare 54 manually annotated images for testing. 
We plot a graph of the displacement against the detection accuracy (more displacement meaning a more deformed surface).  
Figure~\ref{fig:comparison2} shows that our method is able to detect markers with great robustness on deformed surfaces.
Figure~\ref{fig:paper-displacement} validates that our markers outperforms the other methods in detecting the identification of deformed markers.

\begin{figure}[h]
    \vspace{-2mm}
    \centering  \footnotesize
    \begin{subfigure}{.135\textwidth}
    \includegraphics[width=.98\linewidth]{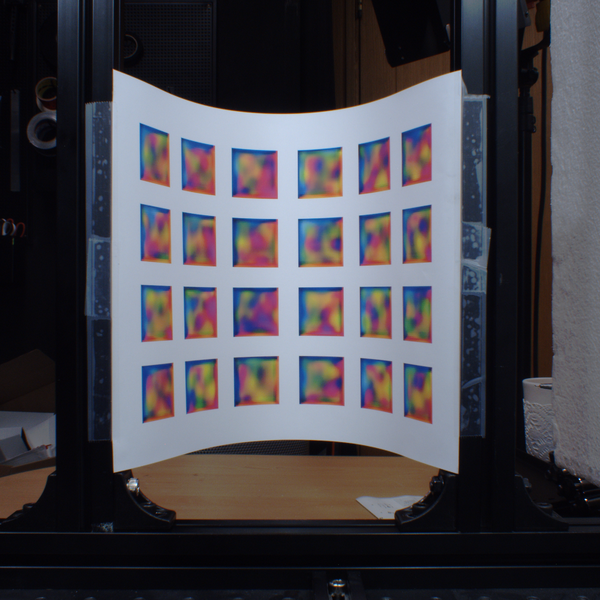}
    \vspace{-1mm}
    \caption{\label{fig:dot-deform-qualitative-convex} Concave}
    \end{subfigure}
    \begin{subfigure}{.135\textwidth}
    \includegraphics[width=.98\linewidth]{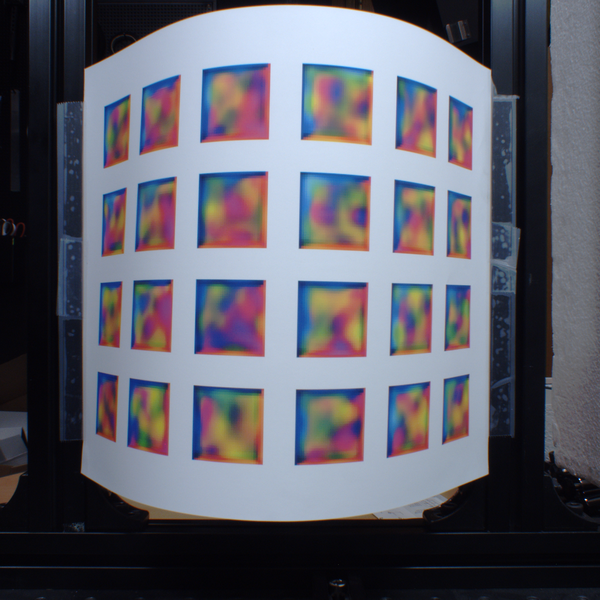}
    \vspace{-1mm}
    \caption{\label{fig:dot-deform-qualitative-concave} Convex}
    \end{subfigure}
	\begin{subfigure}{.18\textwidth}
	\includegraphics[width=.98\linewidth]{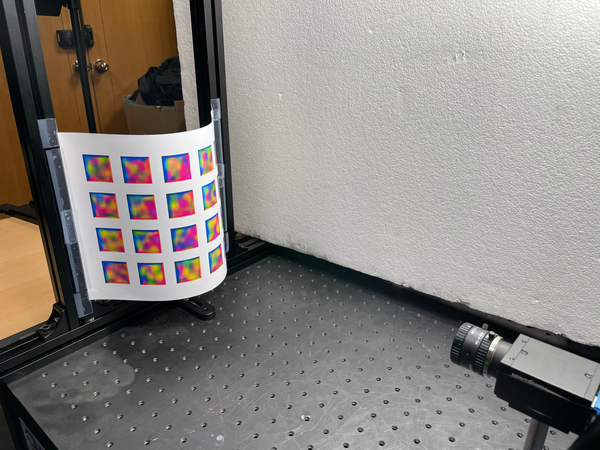}
	\vspace{-1mm}
	\caption{\label{fig:dot-deform-exp-setting} Experimental setup}
	\end{subfigure}
    \caption{
		\label{fig:dot-deform-qualitative} 
		\NEW{Experimental setup for comparing deformable markers.}}
	\vspace{2mm}	\includegraphics[width=0.85\linewidth]{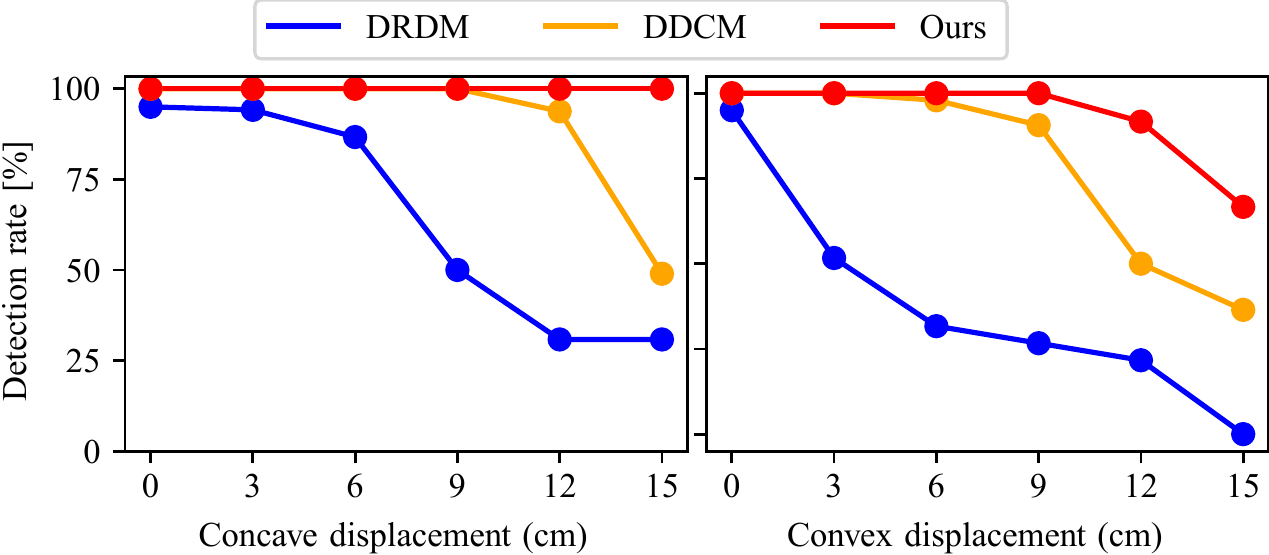}
	\caption{\label{fig:dot-deform-comparison} 
	\NEW{%
	Detection rate against the deformation levels of concave and convex shapes
	compared with DRDM \cite{uchiyama2011deformable}, DDCM \cite{narita2016dynamic}.
	}
	}
	\vspace{-4mm}
\end{figure}

\NEW{%
We compare the detection rate of our learning-based deformable markers with two existing deformable markers based on dot patterns \cite{uchiyama2011deformable,narita2016dynamic}. 
For fairness of comparison, we replicate an experimental setup (Figure~\ref{fig:dot-deform-qualitative}) with the same physical configuration presented in the original paper \cite{narita2016dynamic}.
Then, we compare our detection performance on convex/concave curved surfaces against the performance reported in \cite{narita2016dynamic}. 
Figure~\ref{fig:dot-deform-comparison} validates that our method outperforms these two methods for both cases of deformations.}

\paragraph{\NEW{Real-world flat surface comparison}}
We evaluate the performance of Aruco~\cite{munoz2012aruco} , AprilTag~\cite{wang2016apriltag}, E2ETag~\cite{Peace2020E2ETagAE} and our method on the real-world test dataset. 
Note that we compare our work against methods capable of multi-marker detection and localization since it is crucial for the applications we propose. For example, as the learnable visual markers \cite{grinchuk2016learnable} do not localize and therefore cannot be included in our comparison. 
For fair comparison and to evaluate \NEWB{the} real-world capabilities of each system, we print markers of the selected other works and capture them in the same conditions. 
For a given scene, we capture a set of markers created by four different methods. Among the four shots, we ensure the perfectly matching marker alignment, surroundings and lighting environments to evaluate each method equally.
In the test scenes, we change illumination and camera parameters to cover varied scenarios. We then manually annotate the markers.
We do not apply any augmentations during testing.
We conducted the first experiment of markers placed on planar surfaces. We captured and manually annotated 73 images in total per marker system.
\NEW{Our real-world dataset includes various light sources with correlated color temperatures with a mean of 6,636\,K (standard deviation: 4,542\,K),
 ranging from 1,989\,K to 37,406\,K calculated from sRGB signals in images.}

\begin{figure}[tp]
	\centering  \footnotesize
	\includegraphics[ width=.857\linewidth]{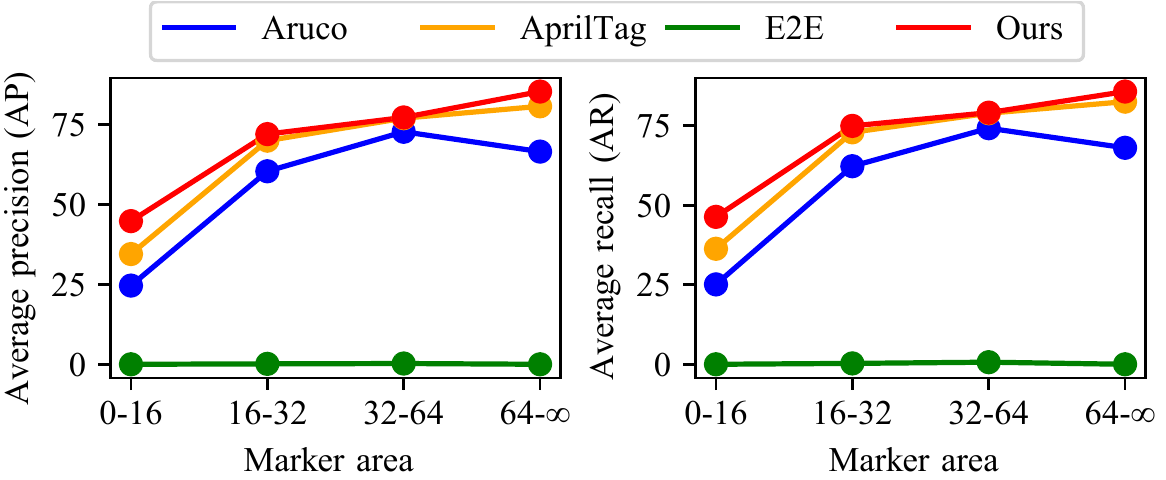}
	\caption{\label{tab:precision_recall} Average precision and recall plots with respect to markers' areas in pixel domain on our flat real\NEWB{-}world dataset. Our method outperforms all the compared methods in terms of all sizes of markers. }
\end{figure}

Figure~\ref{tab:precision_recall} and Table~\ref{tab:main} verify that our tags outperform the others on flat surfaces. We note particularly low accuracy from E2ETag~\cite{Peace2020E2ETagAE}. This is explained by a different layout compared to their training dataset: they place their markers with simple superimposition and constant white boundaries while we place them on larger white paper sheets. In addition, their classification suffers greatly from the presence of multiple E2ETags in a single scene.
	We also ran our method at a lower resolution due to implementation requirements, further harming its detection performance, especially for small markers.
 
\begin{table}[tp]
	\caption{\label{tab:main} 
		We evaluate the performance of three state-of-the-art methods against ours on a real-world test dataset manually annotated. Tags are placed on planar surfaces.
		We use the same spatial resolution of 1216$\times$800 for Aruco, AprilTag, and ours. Due to the implementation limits of E2ETag, we supply input images in a reduced resolution of 640$\times$426. 
		Our method outperforms classical and learned approaches on this planar dataset. The other learned method, E2ETag~\cite{Peace2020E2ETagAE} presents low accuracy when several markers are captured in the same shot due to difficulties in classification.}
	\centering  \footnotesize
	\begin{tabular}{l|c|c|cc|c|c}\hline
		Method & $AP$ & FP-Rate & FPS \\
		\thickhline
		Aruco~\shortcite{munoz2012aruco}       & 50.19 & \red{0.0000} & 31 \\
		AprilTag~\shortcite{wang2016apriltag} & \blue{57.58} & \red{0.0000} & 19 \\
		E2ETag~\shortcite{Peace2020E2ETagAE}   & 00.04 & 0.8625 & 13 \\
		Ours (Aruco markers) 				   & 55.09 & \blue{0.0575} & 29 \\
		Ours (Learned markers)                 & \red{60.84} & \red{0.0000} & 29 \\ \hline
	\end{tabular}
 	\vspace{-3mm}
\end{table}

\section{Applications}
\label{sec:applications}

In this section we propose three main applications that are allowed by our new deformable marker system. Refer to the supplemental video for all results in this section.

\subsection{Structured Light 3D Imaging}

\begin{figure}[tp]
	\centering
	\includegraphics[width=\linewidth]{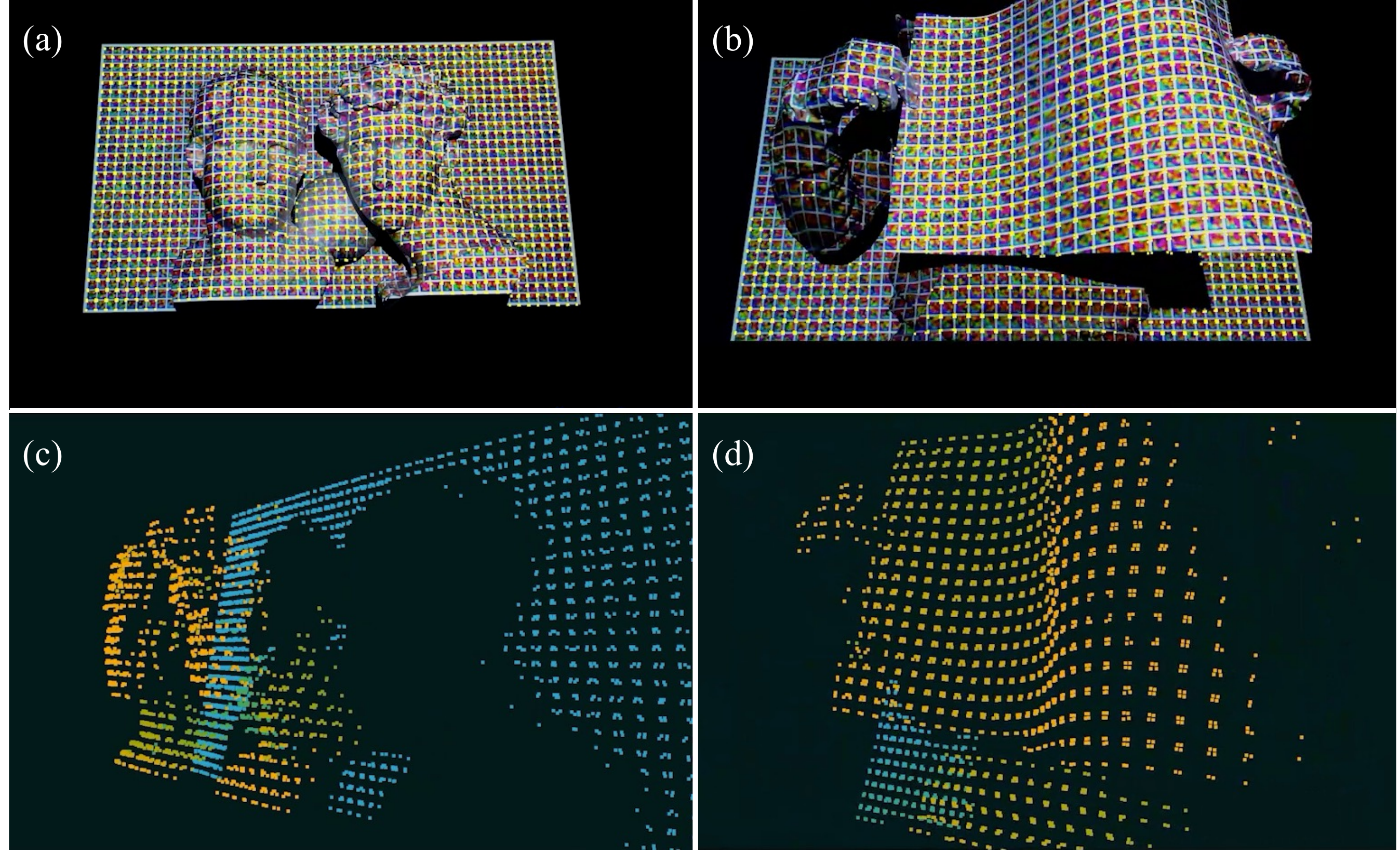}
	\caption{\label{fig:projector-image} Application of estimating depth using multiple deformable markers as structured lighting. 
	(a) and (b) are input images, where  structured light patterns of multiple deformable markers are projected on free-form objects. 
	\NEW{(c) and (d) render point clouds at different viewpoints, which are estimated via stereo using our structured light patterns recognized by our detector.}} 
	\vspace{-1mm}
\end{figure}

For this application, we capture our markers projected densely in a scene by a projector with a stereo camera. 
\NEW{One of the advantages of our marker system is the capability of producing and detecting a very large number of fiducial markers. For structured light 3D scanning, we project a large number of unique markers at each vertex on a 2D grid.}
Using the correspondence matches by the fiducial markers, 
we triangulate the matching corners from both views to estimate 3D geometry. 
\NEW{It is worth noting that the scanning resolution with our marker system is merely limited by
 	the resolution of the camera and projector used, rather than the messaging capability.}
Our results in Figure~\ref{fig:projector-image} show that, without explicit change in training augmentations, 
our markers are able to accurately estimate a large number of 3D points 
\NEW{ even when using a projector on curved surfaces}.

\subsection{Motion Capture of Deformable Surfaces}

\begin{figure}[tp]
	\centering \footnotesize
	\begin{subfigure}{.20\textwidth}
		\includegraphics[trim={1cm, 0cm, 1cm, 0cm}, clip, angle=90, width=0.99\linewidth]{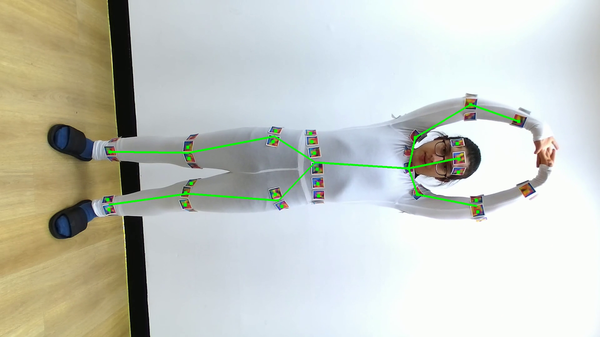}
		\vspace{-1mm}
		\caption{\label{fig:motion_hands} Motion capture frame \#1}
	\end{subfigure}
	\begin{subfigure}{.20\textwidth}
		\includegraphics[trim={1cm, 0cm, 1cm, 0cm}, clip, angle=90, width=0.99\linewidth]{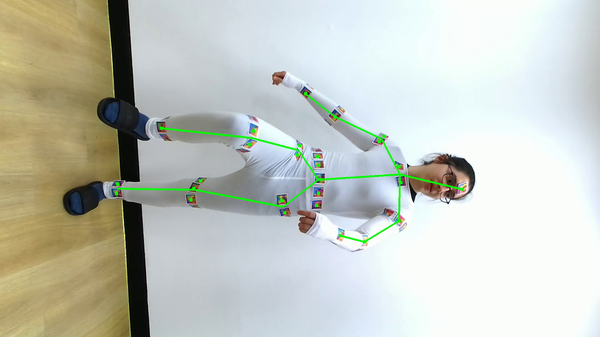}
		\vspace{-1mm}
		\caption{\label{fig:motion_walk} Motion capture frame \#2}
	\end{subfigure} 
	\vspace{-1mm}
	\caption{\label{fig:motion} Application of human motion capture using our markers.}
	\vspace{-1mm}
\end{figure}

Assuming planar placement is a hard constraint, and prevents many additional applications. 
To achieve robustness for deformable markers, we apply strong geometric augmentations 
and we show in Section~\ref{sec:results} that our marker system enables much more robust detection of markers on deformable surfaces.
As an application, we place our markers on  cloth, 
we use our markers like a feature descriptor that we triangulate through a stereo camera. 
For each estimated label points, we calculate the joint points in 3D by averaging the label points in same joint group. 
Figure~\ref{fig:motion} shows that our markers provide successful pose estimation.

\subsection{Augmented Reality Rendering}

\begin{figure}[htpb]
	\centering
	\includegraphics[width=0.9\linewidth]{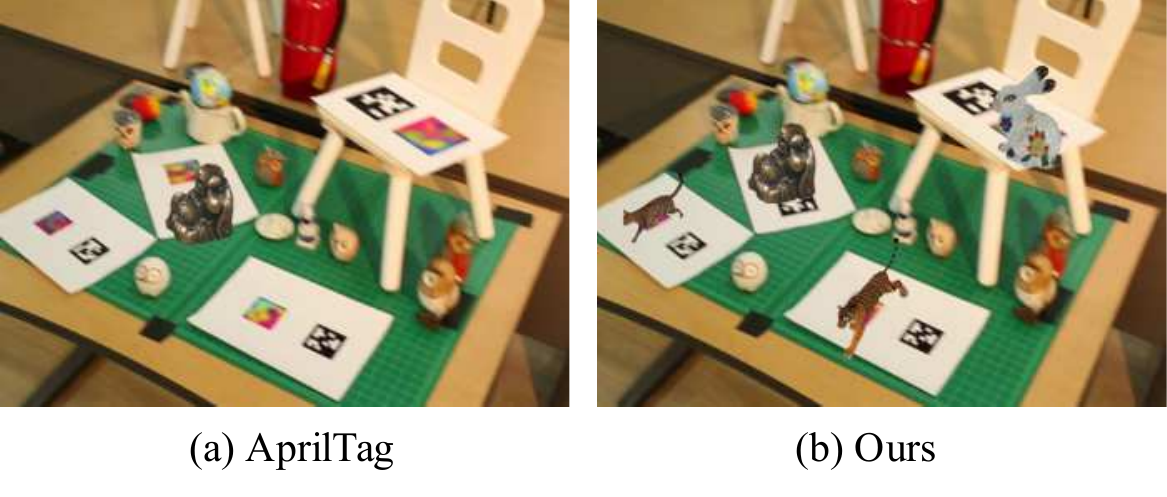}%
	\vspace{-2mm}
	\caption{\label{fig:ar} 
	Application of augmented reality rendering of virtual objects on the detected fiducial markers. 
	(a) shows AR rendering results using AprilTag. In this video frame, fiducial markers are failed to be detected due to strong motion blur by shaking the camera. 
	Only one marker is detected successfully among four markers positions. 
	(b) shows the rendering result using our markers. Our AR rendering does not flicker over frames, showing all four virtual objects successfully. }
	\vspace{0mm}
\end{figure}

When the fiducial marker sizes and camera intrinsic parameters are known, relative marker pose can be estimated. 
While this application is already allowed by existing markers, they fail under extreme imaging conditions such as motion blur and dark illumination. 
Especially in the video setting, where the camera and markers are moving with changing illumination, 
our method provides smooth and consistent detection of the markers. 
Figure~\ref{fig:ar} shows that, in the presence of motion blur, 
the virtual object is placed more reliably thanks to the robust detection performance of our method. 
\NEW{We provide qualitative comparison with AprilTag in the supplementary video. 
	Also, we provide quantitative results of the video in Table~\ref{tab:supp_quant}.}

\begin{table}
	\caption{\label{tab:supp_quant} 
		\NEW{Quantitative detection rates of the supplementary video.}}
	\centering  \footnotesize
	\begin{tabular}{l|c|c|c}\hline
		Experiment & Time & AprilTag~\shortcite{wang2016apriltag} & Ours \\
		\thickhline
		Smooth camera motion  & 01:30-01:38 & 99.94\% & 100.00\% \\
		Dynamic camera motion & 01:38-01:55 & 76.11\% & 99.83\% \\
		Dim illumination      & 01:56-02:13 & 89.81\% & 99.70\% \\
		Halogen illumination  & 02:14-02:19 & 98.40\% & 100.00\% \\
		Deformed markers      & 02:19-02:44 & 44.21\% & 92.58\% \\
		\hline
	\end{tabular}
\end{table}

\section{Discussion}
\label{sec:discussion}

\begin{figure}[pt]
    \vspace{4mm}
    \centering  \footnotesize
    \begin{subfigure}{.14\textwidth}
    \includegraphics[width=.97\linewidth]{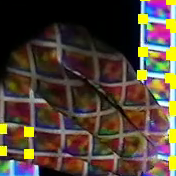}
    \vspace{-1mm}
    \caption{\label{fig:color-failure} Color background}
    \end{subfigure}
    \begin{subfigure}{.14\textwidth}
    \includegraphics[width=.97\linewidth]{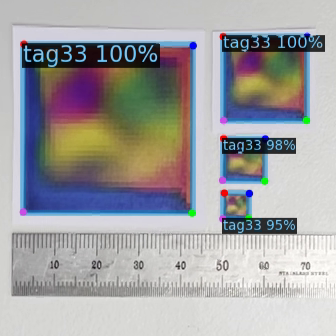}
    \vspace{-1mm}
    \caption{\label{fig:small-marker-failure} Small resolution}
    \end{subfigure}
	\begin{subfigure}{.14\textwidth}
	\includegraphics[width=.97\linewidth]{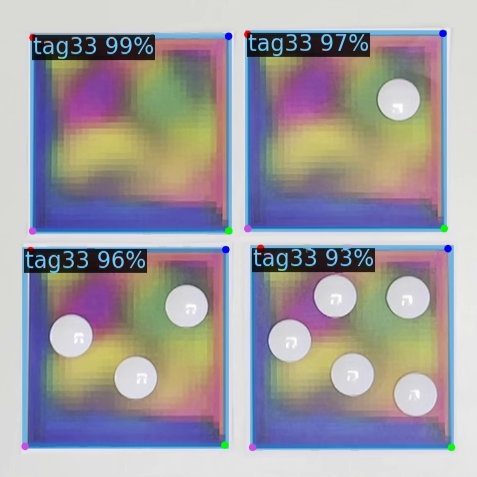}
	\vspace{-1mm}
	\caption{\label{fig:occlusion-failure} Occlusion}
	\end{subfigure}
	\vspace{-1mm}
    \caption{\label{fig:fail-cases}
        \NEW{Failure cases of our method with markers projected on a hand, captured in a very small resolution, and occluded by front objects. The numbers in (b) and (c) indicate the confidence levels of the decoded messages.}}
%    \vspace{-2mm}
\end{figure}

\paragraph{Colored marker}
The capability of \NEWB{producing} a large number of deformable fiducial markers \NEWB{constitutes} the core of our method.
It is possible by using the \NEWB{3-channel colored}
 markers. 
However, there is a tradeoff when using the \NEWB{colored markers}. 
Color appearance and contrast of the captured markers depends on illumination condition. 
Therefore, achieving the robust performance of color fiducial markers requires strong augmentation when training the network. 
Our results verify that our color fiducial marker system can outperform monochromatic systems.
\NEW{%
However, we do not intend robustness against extreme color changes in this work
as color channels are designed to embed information to the markers. 
We observe that markers perform well under the hue-shift levels
within the range of the training hue shift (Figure~\ref{fig:hue_bright_ap}), 
but it could fail when markers are projected or printed on colored surfaces (Figure~\ref{fig:color-failure}).}

\paragraph{Error detection}
Error detection mechanism has been used for verification of decoded messages by checking \NEWB{the} sanity of the inferred messages
with cost of the messaging resource. 
During real-time inference, we reject a fiducial marker's identification that is lower than 80\% match 
of the 36-bit dimension of the binary feature vector. 
As we minimize the domain gap successfully in training encoder/decoder networks 
and achieve the highest AP scores and competitively low \NEWB{false positives},
our system does not include a sophisticated error detection scheme.
\NEW{This reveals itself rarely with small markers with wrongly detected bits (Figure~\ref{fig:small-marker-failure}).}
However, the robustness of marker identification could be improved further by additionally
plugging in an existing error detection mechanism (e.g., minimum distance coding, parity bit, checksum, etc.) to our framework, 
following the current coding systems~\cite{olson2011apriltag,munoz2012aruco}.

\paragraph{Learning with augmentation}
During our training stage, we notice that sometimes the produced data is too difficult or impossible to detect for the algorithm. 
This affects training stability, and sometimes causes exploding loss and gradients. 
To prevent this, we fine-tune our parameters to produce a dataset\NEWB{,} such as minimum and maximum strength values for the augmentations, minimum and maximum marker size in image space for rendering, and specular intensity range.

\paragraph{Specular reflection}
For the specular reflection intensity, we calculate a maximum value that can make brightest point on the board to barely saturate after added to the diffuse component. We also select a random power within the range of zero to the maximum value found. This ensures that the specular component does not fully saturate the surface.
Although these efforts increase stability of training, they could not fully prevent difficult edge cases to be produced. 
We\NEWB{,} therefore\NEWB{,} use adaptive loss functions. 
To achieve this, we keep the running statistics (mean and standard deviation) of the loss value. 
And we clamp the values that are  bigger than $\mu + 3\sigma$ of calculated statistics. 
This prevents \NEWB{explosion of loss or gradients}
when edge cases \NEW{are} produced.

\paragraph{Augmentation of further imaging artifacts}
If certain environmental conditions can be provided as a differentiable function to the network, such as atmospheric haze or underwater, further robustness could be achieved.

\paragraph{Marker occlusion}
Partial occlusion of a fiducial marker is a traditional problem to achieve robust performance in detecting deformable markers.
Currently, a feature that compensates \NEWB{form} some occluded codes within a fiducial marker is 
missing from our current system, 
\NEW{and sometimes degrades the successful performance of localization or decoding (Figure~\ref{fig:occlusion-failure}).}
It is an interesting direction for our future work.

\section{Conclusion}
\label{sec:conclusion}
We have developed \NEW{a novel} deformable fiducial marker method through an end-to-end optimization 
of the marker generator and detector network via photorealistic differentiable rendering. 
Our method outperforms state-of-the-art detection capabilities with both synthetic and real-world image data. 
In particular, our method presents strength in detecting deformed fiducial markers and markers with strong motion blur. 
In addition, the \NEWB{number} of messages that can be embedded in a marker is significantly larger than existing methods. 
It allows us to detect a large number of multiple markers simultaneously in real\NEWB{-}time, 
enabling novel applications such as structured light 3D imaging and human motion capture. 
We anticipate that our method can provide \NEWB{a} more robust performance of augmented reality rendering 
with various real-world imaging scenarios.

\begin{acks}
\NEW{Min H.~Kim acknowledges Samsung Research Funding Center of Samsung Electronics (SRFC-IT2001-04) for developing partial 3D imaging algorithms, 
in addition to the partial support of Korea NRF grant (2019R1A2C3007229), MSIT/IITP of Korea (2017-0-00072), and MSRA.}
\NEWB{We thank Arda Senocak for helpful discussions, 
    Baha Yaldiz for his support on the paper figures and applications, and
    Ayla Yaldiz for making the motion-tracking suit.}
\end{acks}

\bibliographystyle{ACM-Reference-Format}
\bibliography{egbib}

\appendix
\section*{Appendix}
\section{Marker Generator Network Details}
Table~\ref{tab:geneator_network_details} provides details of parameters of our marker generator networks.

\begin{table}[htpb]
		\caption{\label{tab:geneator_network_details} 
		Marker generator architecture details.
		For the convolutions, stride and padding is  set to $(1,  1)$  with zero padding.}
	\centering \footnotesize
	\setlength{\tabcolsep}{3.5pt}
	\begin{tabular}{l|c|c|c}\hline
		Operator &    	Activation 		  &         Input           &  Output shape  \\
		\thickhline 
		Message  	& 			- 		&			 - 				& 		  $36$ 		\\
		FC1      		&	   -     		& 		Message 	  & 		$256$ \\
		Pixel-norm  & 		LReLU   &			FC1			  & 		$256$ \\
		FC2 			& 	LReLU 	   & 	Pixel-norm 		 & 		   $256$ \\
		Reshape 	& 			-	    & 		FC1 			  & 		$4\times4\times16$ \\
		Upsample1   &          -           &    Reshape        & $8\times8\times16$\\
		Conv1 ($3\times3$) & LReLU & Upsample1    & $8\times8\times8$\\
		AdaIn1  &    -                     & Conv1, FC2            & $8\times8\times8$    \\
		Upsample2  &          -           &    AdaIn1        & $16\times16\times8$\\
		Conv2 ($3\times3$) & LReLU & Upsample2    & $16\times16\times6$\\
		AdaIn2  &    -                     & Conv2, FC2            & $16\times16\times6$    \\
		Upsample3   &          -           &    AdaIn2       & $32\times32\times6$\\
		Conv3($3\times3$) & LReLU & Upsample3    & $32\times32\times6$\\
		AdaIn3  &    -                     & Conv3, FC2            & $32\times32\times6$    \\
		ConvLast toRGB ($1\times1$) & Sigmoid & AdaIn3& $32\times32\times3$ \\ \hline
	\end{tabular}
	\end{table}

\section{Corner and Decoder Head Network Details}
Table~\ref{tab:corner_network_details} provides details of parameters of our corner and decoder head networks.

\begin{table}[H]
	\caption{\label{tab:corner_network_details} 
		Corner head and decoder head architecture details.}
	\centering \footnotesize
	\setlength{\tabcolsep}{3.5pt}
	\begin{tabular}{l|c|c|c}\hline
		Operator &    	Activation 		  &         Input           &  Output shape  \\
		\thickhline 
		RoI Align  	& 			- 		&			 - 				& 		  $12\times12\times128$ 		\\
		FC-common &     ReLU		&	   RoI Align       & 		$256\times1$ \\
		FC-resample1  & 		ReLU   &	FC-common		  & 	$128\times1$ \\
		FC-resample2 	& 			-	    & 	FC-resample1 & 		$8\times8\times2$ \\ %
		Decoding sampler 	& 			-	    & 	RoI Align, FC-resample2 & 		$8\times8\times128$ \\ 
		FC3 				& ReLU & Decoding sampler & $512$ \\
		FC4 				& ReLU & FC3 & $256$ \\
		FC-decoder   & - & FC4 & 36 \\
		FC-objectness   & - & FC4 & 1 \\
		\thickhline
		FC-corner1  &     ReLU		&	   FC-common       & 		$128\times1$ \\
		FC-corner2   &         -           &    FC-corner1        & $4\times6$\\ %
		CornerSampler  & - & Stem, FC-corner2   & $4\times8\times8\times64$\\
		Conv3($3\times3$, zero-padding) & ReLU & CornerSampler    & $4\times6\times6\times32$\\
		FC5 & ReLU & Conv3 & $4\times128$\\
		FC6 & ReLU & FC5 & $4\times64$\\
		FC-corner predictor& - & FC6 &$4\times2$\\ \hline
	\end{tabular}
\end{table}

\end{document}